# EGO: a Recursive and Self-Referential Cognitive Architecture for Artificial General Intelligence

**Alberto Mangiante**
Federal University of Alagoas
CNPq Research Group: Enactivism and Digital Systems
Avenida Lourival Melo Mota
57072-970 Maceió, Alagoas
Brazil
MANGIANTE.ALBERTO@GMAIL.COM

**Paolo Totaro**
Federal University of Alagoas
CNPq Research Group: Enactivism and Digital Systems
Avenida Lourival Melo Mota
57072-970 Maceió, Alagoas
Brazil
PAOLOTOTARO1@GMAIL.COM

**Domenico Ninno**
University of Naples Federico II
Via Cintia 21
80126 Naples
Italy
AND
Federal University of Alagoas
CNPq Research Group: Enactivism and Digital Systems
Avenida Lourival Melo Mota
57072-970 Maceió, Alagoas
Brazil
NINNODOMENICO@GMAIL.COM

## Abstract

*Artificial General Intelligence* (AGI) is commonly defined in functional terms as the ability to perform a wide range of cognitive tasks. In contrast, this work adopts a structural perspective, interpreting intelligence as an emergent property of autonomous, self-organizing systems grounded in principles of autopoiesis and embodied cognition. We examine the limitations of current *Large Language Models* (LLMs), including the lack of intrinsic grounding, absence of dynamic self-organization, reliance on external validation, and lack of intrinsic goal-directed behavior. To address these issues, we introduce EGO (*Environment Generative Operator*), a cognitive software architecture based on a formal language (E-language) specifically developed for this project. This language has two particularities. The first is that its expressions, which we call *E-Formulas*, have an intrinsically recursive structure. The second, the most significant, is that it is capable of self-referenciality without reliance on external meta-levels. In EGO, cognition is an emergent phenomenon arising from the conservation of the system's organization. To maintain its organization, EGO continuously modifies its internal structure, thereby realizing Maturana's concept of autopoiesis. The structural changes of which EGO is capable give rise to a dynamic self-organization and an intrinsic goal-directed behavior that are key requirements commonly associated with AGI. A further added value of EGO is that of providing a formal bridge between contemporary artificial intelligence systems and biologically grounded theories of cognition.

## 1. Introduction

Artificial General Intelligence (AGI) is commonly defined as the capacity of an artificial system to exhibit general cognitive abilities across a wide range of contexts (Legg & Hutter, 2007; Goertzel, 2014; Wang, 2019; Sarikaya, 2025; Wang & Sun, 2025). Unlike narrow AI systems, which are designed for specific tasks, AGI is generally associated with adaptability, autonomy, and the ability to operate across multiple domains while continuously modifying its internal organization in response to experience. Recent work has also sought to operationalize different levels of progress toward AGI (Morris et al., 2024).

Two broad perspectives can be distinguished in the AGI literature. The first, largely functionalist, defines intelligence in terms of performance on cognitive tasks, independently of the mechanisms that generate such performance (Legg & Hutter, 2007; Chollet, 2019; Russell & Norvig, 2021; Bostrom, 2014; Wang, 2019; Bubeck et al., 2023). The second, inspired by biology and systems theory, views intelligence as an emergent property of organizational processes, particularly in systems characterized by autonomy, self-production, and structural adaptation (Maturana & Varela, 1980; Varela et al., 1991; Beer, 2000; Di Paolo et al., 2017; Buzsáki, 2019; Muñoz-Cristi, 2025). This distinction is important because a system may reproduce intelligent behavior without implementing the processes that generate cognition itself, a concern raised both in critiques of symbolic computation (Searle, 1980) and in contemporary discussions of large-scale statistical models (Bender et al., 2021).

Large Language Models (LLMs) currently represent the dominant paradigm in artificial intelligence (Brown et al., 2020; Bommasani et al., 2021; Sarikaya, 2025). Despite their remarkable capabilities, they exhibit several limitations from the perspective of biologically grounded theories of cognition. They lack intrinsic grounding, since they operate on statistical regularities rather than direct experience (Harnad, 1990; Bender et al., 2021); they do not continuously modify their structure during operation (Lake et al., 2017; Marcus, 2020); their evaluation criteria remain externally defined, indicating the absence of operational closure (Floridi & Chiriatti, 2020); and they do not possess intrinsic goals or homeostatic mechanisms (Maturana & Varela, 1980; Wang, 2019). As a consequence, current systems are better understood as powerful mechanisms for linguistic simulation than as autonomous cognitive agents.

In parallel, biologically inspired approaches have emphasized the role of embodiment, sensorimotor interaction, self-organization, and structural coupling in the emergence of cognition (Brooks, 1991; Varela et al., 1991; Beer, 2000; Di Paolo et al., 2017). Within these frameworks, cognition is not viewed as the manipulation of predefined representations, but as the outcome of recursive processes that preserve the identity of the system while allowing structural adaptation. This perspective is reinforced by the “inside-out” approach proposed by Buzsáki (2019), according to which cognition emerges primarily from endogenous dynamics rather than from the passive processing of external stimuli. The present work adopts this organizational perspective. We introduce the Environment Generative Operator (EGO), a framework for Artificial General Intelligence composed of a set of interacting algorithms and software modules designed to implement core principles derived from autopoietic theories of cognition. The complete EGO software suite, together with its technical documentation and explanatory notes, is publicly available at https://sourceforge.net/projects/ego-agi/.

The EGO architecture is based on the hypothesis that intelligence emerges from the interaction between recursive self-organization, intrinsic grounding, operational closure, and goal-directed behavior. Grounding arises through the categorization of sensory stimuli; operational closure is implemented through a self-referential formal language; and goal-directedness is formalized as the preservation of homeostatic equilibrium. Within this framework, cognition is not predefined but emerges progressively through a transition from a precognitive phase, characterized by the clustering of sensory inputs, to a cognitive phase in which representations become self-referential and capable of supporting adaptive behavior.

The main contribution of this work is twofold. First, it proposes a formal architecture that operationalizes key concepts from autopoietic theories of cognition, including the distinction between organization and structure, operational closure, structural coupling, and homeostatic regulation. Second, it demonstrates the feasibility of implementing these principles within a computational system capable of generating adaptive behavior from recursive and self-referential mechanisms. In this sense, EGO is intended not merely as a new architecture for AGI, but as a reformulation of the criteria by which intelligence itself may be modeled in artificial systems.

The remainder of the paper introduces the E-language, the Self-Referentiality Theorem, the recursive mechanisms and the corresponding algorithms that constitute the computational core of the proposed architecture.

## 2. The E-language as the Operational Domain of EGO

EGO is intended to operate under the conditions required of an autopoietic system. This means that the products of its activity must remain available for subsequent operations without requiring translation into an external representational language. The algorithm therefore requires an object language in which every computational result can itself become the object of further recursive computation. The E-language has been designed to satisfy this requirement.

Unlike conventional symbolic languages, whose expressions often require additional representational levels in order to describe relations among themselves, the E-language allows relations to be expressed by formulas belonging to the language itself. Consequently, EGO can recursively operate not only on symbolic structures but also on the relations that emerge among those structures. This property provides the symbolic basis of the algorithm's operational closure.

The language is generated from a single primitive symbol through a recursive syntax formally specified in Appendix A. Every well-formed expression of the language, called an *E-formula*, belongs to the same symbolic domain regardless of the recursive level at which it is generated. Thus, expressions such as **(0)**, **((0))**, and **(0, (0, (0, (0))))** are all valid E-formulas, differing only in the complexity of their recursive organization.

A distinctive feature of the E-language is that every E-formula admits two complementary interpretations. It can be interpreted as a finite set recursively generated from the primitive symbol **0**, but the same expression can also be interpreted as a proposition of a two-valued logic. This dual interpretation allows EGO to move continuously between structural representation and logical evaluation without leaving the symbolic domain in which its computations take place. In other words, our E-language is *homoiconic.*

The key property required by EGO is provided by the *self-reference theorem*, formally proved in Appendix A. The theorem shows that, given any two E-formulas $X$ and $Y$, it is always possible to construct another E-formula, denoted as $(X = Y)$, whose logical value evaluates whether the two original expressions are equal when interpreted as sets. Since every set-theoretical relation can ultimately be expressed in terms of equality, the same recursive construction extends naturally to all set relations. Consequently, relations among E-formulas become E-formulas themselves.

Figure 1 illustrates the recursive construction of the E-formula $(X = Y)$, providing an intuitive visualization of the procedure described by the theorem. Rather than comparing two expressions as linear strings, the construction recursively compares the corresponding levels of their rooted-tree representations. Whenever corresponding nodes are themselves compound expressions, the comparison continues recursively; when the recursion reaches the primitive level, logical equivalence replaces further decomposition. The result of the entire evaluation is again an E-formula. Thus, the outcome of evaluating a relation remains within the same symbolic domain in which subsequent computations will occur.

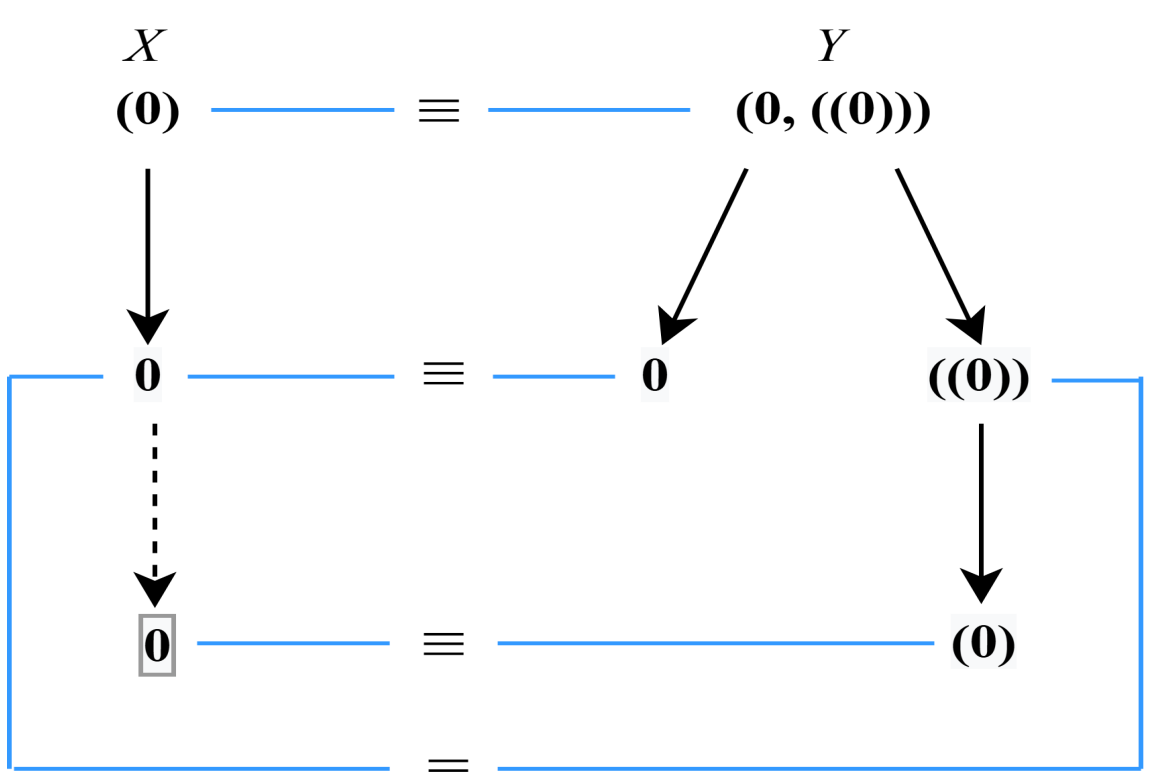

**Figure 1.** Illustration of the recursive construction of the equality evaluator **(**X = Y**)**. The recursive procedure compares the corresponding nodes of the rooted-tree representations of two E-formulas, generating an E-formula that evaluates their set-theoretical equality. Because the result is itself an E-formula, it can immediately become the input of subsequent recursive computations.

The importance of this result for EGO goes beyond the evaluation of equality. We shall refer to $(X = Y)$ as the *equality evaluator*, since it is itself an E-formula evaluating whether the two E-formulas $X$ and $Y$ are equal when interpreted as sets. Starting from the equality evaluator, evaluators for every set relation and operation can be recursively constructed (see Appendix A). More generally, we denote by $(X \text{ R } Y)$ an E-formula that evaluates whether the relation R, expressed in the metalanguage, holds between the E-formulas $X$ and $Y$.

The self-reference theorem therefore transforms recursive computation from a linear process into a hierarchical one. Suppose, for example, that two E-formulas $X$ and $Y$ represent endogenous or exogenous perturbations affecting the system. Through the equality evaluator and its generalization to relation evaluators, the algorithm can construct an E-formula **(**$X$ R $Y$**)** expressing whether a particular relation R holds between those perturbations. That newly generated EF can itself become the argument of subsequent recursive operations. Thus, given two relation evaluators $R_1$ = **(**$X$ $R_1$ $Y$**)** and $R_2$ = **(**$X$ $R_2$ $Y$**)**, EGO can generate another EF $R_3$ = **(**$R_1$ $R_3$ $R_2$**)** expressing whether the relation $R_3$ holds between the previous relations.

The hierarchy generated in this way has no predetermined upper bound. Relations become objects of further relations without leaving the language in which they are expressed. Because the organization of an autopoietic system consists precisely of relations among its components, this property allows EGO to represent, evaluate, and recursively preserve aspects of its own organization entirely within its operational domain, without requiring any external representational level.

## 3. EGO initialization

In the previous section, we introduced the E-language and showed that it supports both recursion and self-reference, all without ever having to step outside the language itself. Indeed, it can be shown that the E-language is Turing-complete, meaning that it is capable of simulating a Turing machine. Consequently, at least in principle, it would be possible to implement the entire EGO system within the E-language itself. However, we did not pursue this approach because our intention was—and still is—to demonstrate, in the simplest and most readable way possible, EGO's ability to realize a form of Artificial General Intelligence by operationalizing the theory of cognition developed by H. Maturana. With this objective in mind, we used E-language as the object language and Java, running on a Spring Boot architecture, as the "interpreter" for E-language string manipulation operations.

EGO is made up of two types of modules, the *Environment* and *Individual*. In some sense, the Individuals behave as living agents while the Environment simulates the external world. From a software point of view, both the Environment and the Individual modules use Apache Kafka for all the communication, MongoDB as a memory system, and the K-means model, as implemented in Apache Spark, for the categorization of "sensorial stimuli". The overall software architecture of EGO is shown in Figure 2.

We begin with a coarse-grained description of how EGO operates, postponing the discussion of the underlying mechanisms and implementation details to the Appendices. Each individual shown in Figure 2 is a software entity characterized by its own organization and structure, both expressed in terms of E-formulas. The meanings we assign to the concepts of *organization* and *structure* correspond exactly to those introduced by Humberto Maturana. Recall that, while an individual's internal structure may change in response to stimuli originating from the environment, its organization must remain invariant over time. In biology, this property is known as *homeostasis*. More specifically,

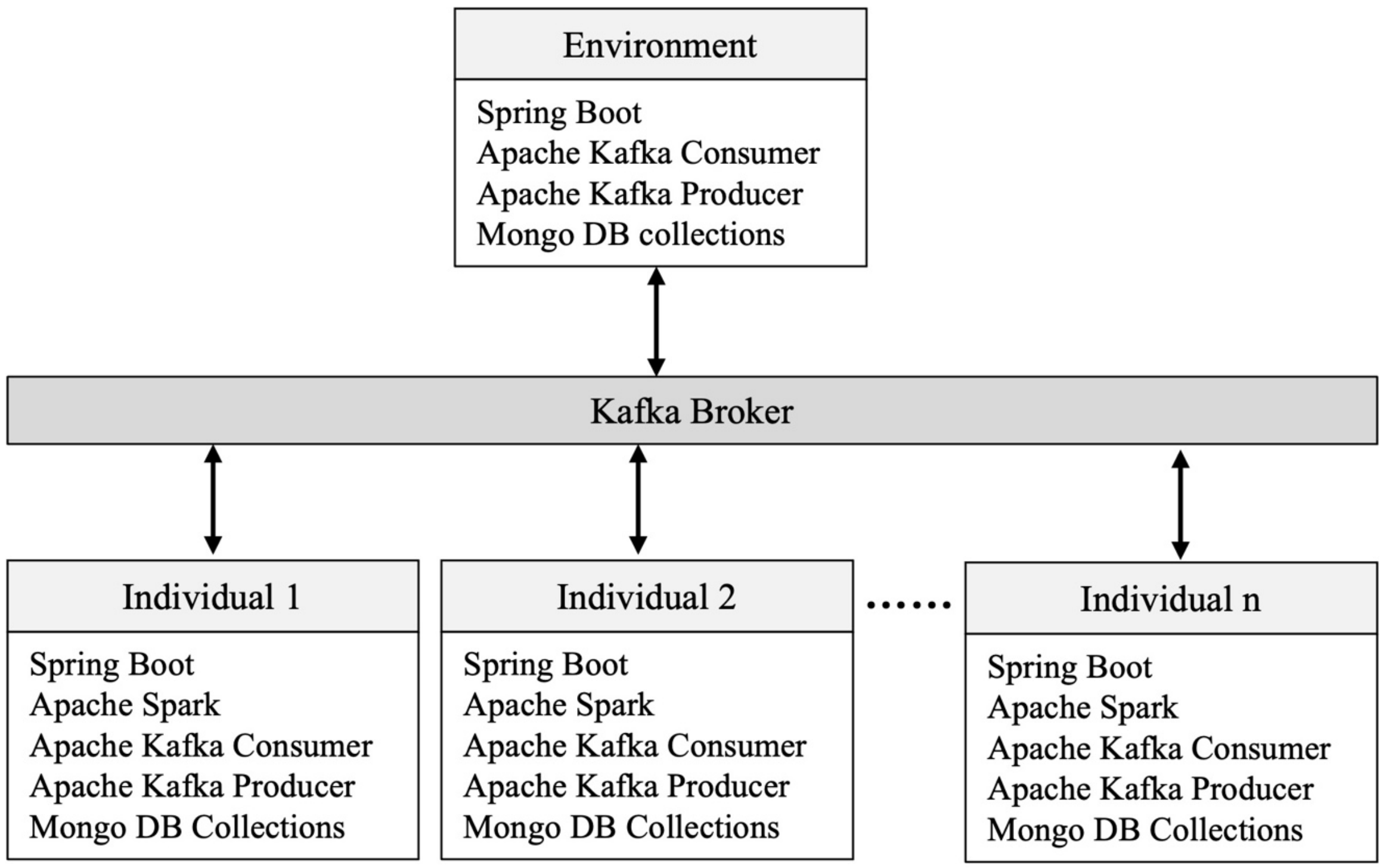

**Figure 2** Architectural layout of EGO. The system comprises two main modules type, Environment and Individual which communicate asynchronously through a Kafka broker. Up to n Individual instances can be deployed in parallel, each consuming events from the whenever an environmental stimulus perturbs the system, the individual attempts to modify its structure in such a way as to restore its organization to its original state. If this process fails—that is, if the organization becomes irreversibly disrupted—the individual ceases to exist. Within Maturana's theory, the drive to preserve organization and maintain viability constitutes the fundamental source of autopoietic cognition.

When a run of EGO starts, the first step is the creation of the environment in which the individuals will subsequently operate. At this stage, no individuals have yet been generated. The entire initialization of the environment is carried out by the Environment module. A clarification is appropriate here. The objective of EGO is not to reproduce the physical or chemical properties of a realistic environment. Instead, we assume that, from the perspective of cognition, the essential complexity of an environment can be abstracted into two fundamental points: a collection of elementary components and an internal organization relating these components. In the computational implementation, the elementary components are represented as bit sequences, whereas the internal organization is represented by the way these sequences are grouped around a set of centroids, optionally according to a specified probability distribution. This classification process is performed by the K-means modules. Consequently, at the beginning of each run, the Environment module generates a collection of clusters of bit sequences, each characterized by its own centroid. This clustered structure constitutes the environment in which the individuals are subsequently generated and within which all their interactions take place. Since centroids are computed as averages over the elements belonging to a cluster, they can also be interpreted as *Common Aspect*, in the sense that they represent the features shared by the corresponding cluster members. The individual elements belonging to a cluster will be referred to as *Instances*.

As will be shown later, both Instances and centroids are subsequently converted into E-formulas. This conversion from a sequence of bits to an E-formula is done with the algorithm illustrated in Appendix B. To distinguish the binary representation from the E-formula representation, we will use the adjective *raw* when referring to the binary form. Thus, we shall speak of *Raw Instances* and *Raw Common Aspects*. The Environment module operates exclusively on binary representations and therefore manipulates only Raw Instances and Raw Common Aspect.

*Raw Categories* are defined as one-dimensional arrays of Raw Common Aspect. Categories also possess their own instances. A simple example may help clarify these concepts. Suppose we have three Raw Common Aspects whose decimal values are 100, 200, and 300. We can construct a Raw Category represented by the array [100, 100, 200, 300] (we will see later how, in general, these constructions are made). Now consider the array [98, 102, 205, 304] as a Raw Instance, where 98 and 102 belong to the cluster whose centroid is 100, 205 belongs to the cluster centered at 200, and 304 belongs to the cluster centered at 300. The array [98, 102, 205, 304] is therefore an instance of the category [100, 100, 200, 300].

Starting from Raw Categories, the Environment module labels some of the instances as raw perturbation instances, others as raw common-aspect instances, and still others as raw instances of emergent outcomes. In this way, the environment defines a collection of potential transformations in which a perturbation X acting on an existing structure Y gives rise to new emergent states Z.

Once the environment creation phase has been completed, the Environment module proceeds with the creation of the individuals, our living agents. The Environment module uses K-means to train individuals to recognize and classify the stimuli generated by the environment itself. These stimuli are selected randomly from the clusters whose centroids were established during the initialization phase. Based on the received stimuli, each individual constructs its own set of centroids. In other words, the individuals' "sensory systems" are trained to recognize incoming stimuli and assign each of them to the appropriate centroid. At this stage, however, we are still in what we call the “precognitive” phase. The individual is not yet aware of common aspects as such; rather, it possesses only its own internally generated version of the Raw Common Aspects obtained through the K-means learning process. The purpose of training is therefore to establish the individual's precognitive capabilities, that is, to configure the sensory system responsible for decoding and pre-categorizing environmental stimuli. At this point, the individual has not yet transformed the centroids produced by K-means into E-formulas and consequently has not yet attributed any cognitive meaning to them. The centroids remain merely statistical regularities extracted from sensory experience, rather than elements of an explicit cognitive representation.

For individuals as well, Raw Common Aspects are grouped into Raw Categories. The structure of each individual is therefore represented by Raw Instances belonging to a subset of these Raw Categories. A simple example may help clarify this process. Suppose that the data stream has produced the raw categories A, B, C, D, E, F, G, H, and I, and that the current run contains three individuals. The first individual may be assigned, for instance, categories A, B, and C; the second categories D, E, and F; and the third categories G, H, and I. The structure of an individual consists of raw instances, that is, arrays of values, each of which belongs to at least one cluster centroid. Consider, for example, the first individual. Suppose that it has been assigned Raw Category A, whose centroids are [100, 100, 200, 300]. Further assume that the individual's initial structure must contain two instances of this category. The environment module may then provide, for example, the two raw instances [98, 102, 205, 304] and [101, 105, 198, 308]. When the first instance, [98, 102, 205, 304], is received, the individual's K-means classifier determines the corresponding centroids and therefore the associated class values. In this case, the classifier identifies the centroid sequence [100, 100, 200, 300], thereby recognizing the received instance as belonging to raw category A. It is worth emphasizing once again that, at this stage, the individual is capable of recognizing similarities and classifying incoming stimuli, but the resulting categories remain purely precognitive. They are statistical classifications derived from sensory experience and have not yet been transformed into E-formulas, the representational format through which cognitive processes are subsequently established within EGO. At this point, the binary values are finally transformed into E-formulas according to the algorithm described in Appendix B. It is important to stress that it is precisely at this point that the cognitive phase begins, in which, for instance, E-formulas give rise to other E-formulas, namely the categories. Appendices C and D provide examples illustrating in some detail how categories are constructed.

As this section has introduced a considerable number of concepts, it is worth providing a brief summary before proceeding further. At the end of the initialization and precognitive phases, each individual possesses:

a) an organization composed of a set of categories derived from K-means centroids; and

b) a structure consisting of the instances associated with those categories.

The categories define the individual's organization, whereas their instances constitute its structure. The instances belonging to the categories that constitute the individual's initial organization are referred to as *Internal States*.

The organizational rule governing each individual—that is, the rule whose preservation ensures the individual's continued existence—is defined as follows. Let M denote the total number of categories in the individual's organization. For each category j, the number Kj of Internal States associated with that category must remain constant. As long as this condition is satisfied, the individual's organization is preserved; if it is violated, the organization is altered and the individual's viability is threatened.

## 4. The living phase

Once the initialization and precognitive phases have been completed, the Environment module begins sending to each individual sequences of binary values generated by applying small random deviations from the centroids of the raw classifications of the perturbations. These sequences constitute the perturbations that may alter the number *K* of Internal States associated with each category. Each individual classifies incoming perturbations through K-means and transforms them into E-formulas. If these E-formulas are not instances of any category belonging to the individual's organization, they are referred to as *Emerging Results*.

Given the significance of this point, it is worth exploring it further. Perturbations arrive as vectors of binary numbers, which are subsequently classified and converted into E-formulas. Within the system, they are treated as perturbations rather than emergent outcomes, because they constitute a flow that leaves no instances behind in the structure.

The instances that do emerge are those of the Emergent Results, whereas certain instances of Y disappear—and this is precisely why (X,Y) = Z, since Z emerges from instances of X and Y. X does not become part of the structure, simply because its instances immediately couple with Y and transform it. They are, in effect, perturbations: they exist, are perceived by the sensory system, and are therefore classified at a conscious level, yet the instances of X do not enter the structure; rather, they modify the instances of Y, giving rise to those of Z. I hope this makes things clearer.

The appearance of Emerging Results may alter the distribution of Internal States across categories. As a consequence, the prescribed number K of Internal States associated with one or more categories may no longer be satisfied, thereby compromising the individual's organization. In other words, the individual may lose its homeostatic equilibrium because some categories no longer contain the required number of Internal State instances. When this occurs, EGO attempts to reconstruct the missing Internal States. To do so, it first searches for Common Aspects among the Emerging States and groups them into new categories, referred to as *Emerging Categories*. These categories do not belong to the individual's original organization but provide the raw material from which replacement Internal States can be generated.

By manipulating these newly generated categories, EGO attempts to reconstruct the missing Internal States, that is, new E-formulas that are instances of the organizational categories. The objective is to restore the original values of K and thereby recover the individual's homeostatic equilibrium. To achieve this goal, an EGO individual may also interact with other individuals, exchanging states that each considers spurious with respect to its own organization. In this way, individuals can draw not only on their own resources but also on resources originating from other individuals that are themselves engaged in restoring their organizational integrity. Through this cooperative process, each individual attempts to preserve its organization despite the structural changes induced by environmental perturbations.

It is important to note that, at the end of each cycle of interaction with the environment, every individual preserves its organization while potentially modifying its structure. Such structural changes may involve not only the generation of new categories but also the emergence of new relationships among categories. Consequently, although the organizational constraints that define the individual's identity remain invariant, the structure through which those constraints are realized can evolve continuously in response to environmental perturbations.

It is worth emphasizing once again that, during the initialization and precognitive phases, the *Environment* module creates the environment within which individuals will subsequently operate. In this sense, it simulates a natural environment that provides individuals with their organization—one might regard it as the equivalent of a genetic code—and that contains both exogenous shocks (perturbations) and the reactions to those shocks (emergent outcomes). At this stage, individuals possess no knowledge of the environment. Their knowledge of it will emerge only through subsequent interactions mediated by their computational mechanisms based on E-formulas. Everything generated by the *Environment* module, including the reactions it defines, is represented exclusively as arrays of binary values and never as E-formulas. Consequently, this phase of a run is entirely independent of EGO's cognitive mechanisms. Its sole purpose is to provide a simulated environment within which cognition can later emerge and operate. The generation of environmental structures, perturbations, and emergent outcomes takes place outside the cognitive domain of the individuals and is therefore completely decoupled from their internal processing.

From a software perspective, the interaction between individuals and the environment is mediated through Kafka topics, which may be viewed as the functional equivalent of the individuals' sensory systems. Kafka topics are also used for communication among individuals. Such communication is likewise carried out through arrays of binary values which, upon reception, are translated back into categories and instances and ultimately into E-formulas. In this way, both environmental perception and social interaction are grounded in the same representational substrate while remaining cognitively meaningful only after the binary information has been transformed into E-formulas.

### 4.1 Archetypes

Before describing the Manipulation and Behaviour modules, which are responsible for EGO's cognitive activity, we must first examine the structure of E-formulas. As can be readily inferred from the recursive definition presented in Section 2, the length of E-formulas grows very rapidly. However, recursion also provides a significant advantage: it naturally induces a tree structure that can be exploited to accelerate manipulation operations and, consequently, computation itself.

To address the computational challenges associated with large E-formulas, EGO employs a data structure called an *Archetype*. Archetypes can be viewed as compact hierarchical representations of E-formulas. Their purpose is to preserve the recursive structure and semantic content of E-formulas while avoiding the computational costs associated with manipulating very large expressions directly. In practice, the E-formulas manipulated by EGO are represented through archetypes. In the remainder of this section, we focus on category archetypes, since they provide the most intuitive illustration of the underlying representation.

An archetype consists of two fields, which we call *name* and *meaning*. For example, the archetype associated with a category B can be written as follows:

(N_Cat(B), Meaning_Cat(B))

where N_Cat(B) denotes the name of category B, while Meaning_Cat(B) represents its meaning. It is useful to note at the outset that the name acts as a label, whereas the meaning may either refer to another archetype or correspond to a terminal leaf of the tree. A concrete example is provided by the category archetype shown in Figure 3. Archetypes are classified into different types, a topic that will be discussed shortly. For the moment, it is sufficient to observe that the name identifies a category, while the meaning encodes its associated tree structure.

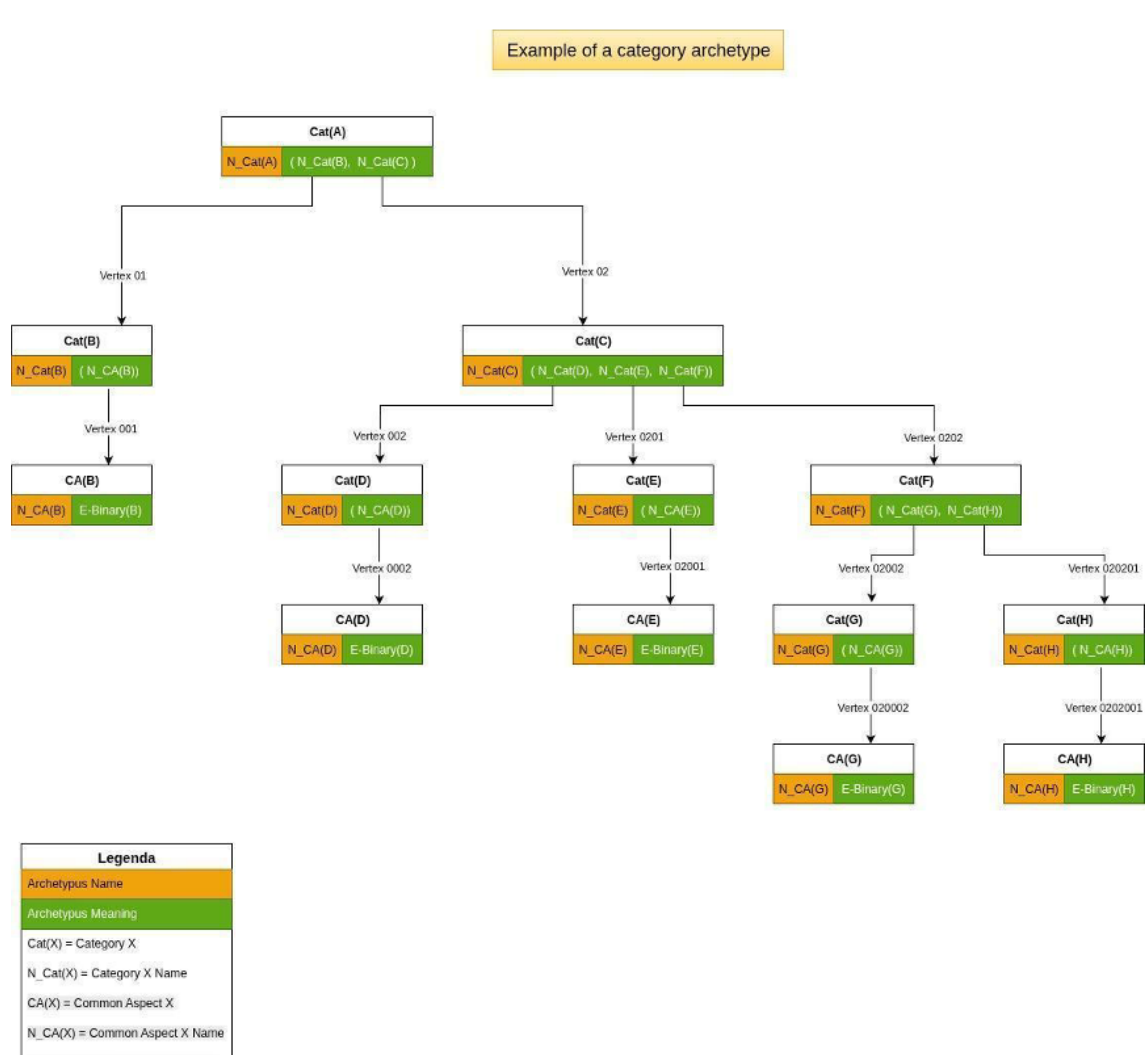

**Figure 3** An example of a category archetype.

As shown in Figure 3, the parent archetype Cat(A) consists of a name, N_Cat(A), and a meaning. The meaning, in turn, contains the names N_Cat(B) and N_Cat(C) of two other archetypes, which constitute the elements of category Cat(A).

If we move down to the second level of the tree, we encounter the categories Cat(B) and Cat(C). The branch labeled N_Cat(B) leads to Cat(B), while the branch labeled N_Cat(C) leads to Cat(C). The names of archetypes therefore act as labels identifying the branch to be followed in the next step of the traversal.

Let us now continue the examination of the second level. The meaning of Cat(B) contains not the name of another category, but the name of a Common Aspect, denoted by N_CA(B). This name, in turn, refers to an E-formula, which corresponds to a terminal leaf of the tree. The leaves of the tree are always Common Aspects, that is, the centroids generated by the K-means algorithm. The remaining branches are interpreted according to exactly the same recursive principle.

From this example, it becomes clear how the archetype of a category can be used to construct a tree structure leading from the initial category down to its leaves, which correspond to Common Aspects. Indeed, the archetype names appearing within the meaning of a category are themselves archetypes representing categories. In turn, the elements of these categories may be either other categories or Common Aspects. Whenever the meaning of a category contains a Common Aspect—that is, a centroid generated by K-means—a leaf of the tree has been reached. Otherwise, it is possible to continue descending through the hierarchy. This hierarchical structure will play a fundamental role in the definition of Manipulation, the operation through which an individual restores its structure and thereby preserves its viability. The nodes of the tree defined by category archetypes are assigned the same encoding used for the nodes of E-formulas, whose construction algorithm is presented in Appendix E.

The relationship between category archetypes and E-formulas deserves special attention. It can be shown that the archetype of a category can be replaced by a single E-formula containing the meanings of all its Common Aspects. In other words, archetypes and E-formulas carry the same information, but in different forms. The archetype provides a compact hierarchical representation, whereas the corresponding E-formula represents the same content as a single recursive expression. However, as already noted, such an E-formula may become extremely large and therefore computationally expensive to manipulate. This is precisely why archetypes are employed. They preserve the semantic content and recursive organization of E-formulas while providing a compact representation that makes manipulation computationally tractable.

EGO employs five types of archetypes:

1. **Perception:** This archetype records the exogenous perturbations affecting an individual. Its meaning contains the E-binary value representing the perceived stimulus.
2. **Common_Aspect:** This archetype represents a common feature. As discussed earlier, a Common Aspect corresponds to the centroid of a cluster of points. Accordingly, the meaning of this archetype contains the centroid value expressed as an E-binary.
3. **Interpretation:** This archetype represents the interpretation of an exogenous perturbation that alters a category and changes the number of certain Internal States, transforming them into instances of categories referred to here as Emergent Outcomes. The meaning of the archetype includes the names of the Perturbation, Internal State, and Emergent Outcome categories involved in the transformation, together with the number of Perturbations applied, the number of Internal States transformed, and the number of Emergent Outcome instances generated.
4. **Manipulation:** This archetype contains a set of instructions for constructing a new instance of an Internal State by using instances of categories that do not belong to the current structure. Its meaning contains the result of a self-programming activity aimed at generating an action plan for restoring the missing instances of a given category. Since manipulation is a complex process, it will not be discussed further at this point. Both the process itself and the resulting archetype will be examined in detail in the dedicated section.
5. **Category:** This is the archetype described in the previous example.

As indicated above, more details on how EGO generates categories can be seen following the examples of Appendices B and C.

### 4.2 Manipulation and Behaviour

When the number of Internal State instances associated with each category matches the values prescribed by the organizational rule, the individual is said to be in a state of homeostatic equilibrium. A violation of these rules alters the distribution of Internal States across categories and thereby disrupts homeostasis. To restore its original organization, the individual must react by reconstructing the missing Internal States.

This recovery process is divided into two phases. The first, called *Manipulation*, consists in determining a plan for restoring the organizational constraints. The second, called *Behaviour*, consists in executing that plan. Manipulation therefore specifies what must be done, whereas Behaviour implements the corresponding actions. Together, these processes enable the individual to preserve its organization despite structural changes induced by environmental perturbations. To understand how Manipulation operates, we must once again consider the tree structure defined by category archetypes. As discussed previously, the meaning of a category contains either references to other categories or, at the leaves of the tree, Common Aspects generated by the K-means algorithm. The recursive organization of these references makes it possible to represent each category as a hierarchical tree whose internal nodes correspond to categories and whose leaves correspond to Common Aspects. The nodes of these trees are assigned the same encoding used for E-formulas. Moreover, the node-indexing scheme introduced in Appendix F identifies not only the position of a node within the tree but also its lineage relationships. This information is essential for Manipulation because it allows the system to efficiently locate structurally corresponding elements across different categories.

Given these premises, the operation of Manipulation can be readily understood. Suppose that we wish to reconstruct an Internal State belonging to category Yj using instances of Emergent Results. Further suppose that instances of category Zi are available. If a sub-element of Zi located at node, say abc, belongs to the same category as a sub-element of Yj located at node, say efg, then the former can be used to reconstruct the latter. The process continues by examining all available Zi categories until sufficient elements have been identified to reconstruct the entire structure of Yj. The search is not limited to the direct elements of Yj but extends recursively to their descendants. Once the lower-level elements have been reconstructed, the higher-level elements can in turn be rebuilt, ultimately allowing the reconstruction of the entire category. EGO determines whether a node of Yj corresponds to a node of Zi by applying Theorem 2 of E-language to the names of the associated archetypes. This requires a systematic comparison of the nodes belonging to the two category trees. If a node of Zi belongs to the intersection of Yj and Zi but is itself a descendant of a node already identified as part of the intersection, it is discarded. Likewise, descendants of Yj whose parent nodes have already been selected are excluded from further consideration. A concrete example of the Manipulation process is provided in Appendix E.

## 5. Discussion and conclusion

In EGO, individuals are constructed whose perception of the world is mediated by a sensory system. The mechanism responsible for generating the bit strings entering the system can be viewed as analogous to a camera, a microphone, a temperature sensor, or any other sensing device. These bit strings are subsequently grouped into clusters by the K-means algorithm, providing the basis for category formation. Since the input data may originate from a camera, a microphone, a thermometer, or any other sensory device, the proposed mechanism is inherently domain-independent and can be adapted to virtually any sensory modality. This universality is consistent with Andrew Ng's hypothesis that a common learning algorithm underlies the functioning of different cortical areas, independently of the sensory modality they process.

We know that biological organisms possess multiple sensory modalities, including vision, touch, smell, and hearing, as well as additional modalities such as proprioception, the vestibular sense, and interoception. For this reason, the next stage of development will focus on creating an individual equipped with a multisensory system rather than the single sensory modality employed in the current version. In such a setting, categories and their corresponding instances—that is, the objects constituting the individual's world and their abstract representations—will be generated from common aspects extracted from one or more sensory modalities. This extension will make it possible to investigate how information originating from different sensory channels can be integrated into unified cognitive structures and how such structures contribute to the emergence of a coherent representation of the environment.

The encoding and categorization mechanism is not only universal—that is, applicable to all sensory modalities—but is also based exclusively on recursion and on E-language, a language that enables operations to be performed on its own expressions without requiring a higher-level supervisory language. This feature is consistent with current views of brain organization, according to which no separate "brain supervising the brain" exists; rather, cognitive processes emerge from the interactions among neurons themselves. At the same time, it greatly simplifies the architecture of the system. In the case of E-individuals, both knowledge and sensory processing can scale without fundamental limitations in terms of:

• **The number of sensory modalities.** The mechanism responsible for generating categories and category instances is universal and can therefore be applied to any sensory modality.

• **The formation of cross-modal categories.** Categories may contain elements originating from different sensory systems without raising representational compatibility issues. In an individual, all category elements—including categories derived from different sensory modalities—are represented as E-formulas and are therefore processed according to the same linguistic rules. By contrast, in architectures where one language supervises another, the rules governing category representation may differ across sensory modalities, either because the supervised and supervisory languages are different or because the relationship between them necessarily introduces additional representational constraints.

• **The number of generated categories.** The growth of categorizations does not lead to an uncontrolled computational explosion because of the use of archetypes. Archetypes, which are themselves E-formulas endowed with a name and a meaning, provide a compact representation of knowledge structures. Once again, the solution is achieved entirely through recursion and mechanisms internal to E-language, without requiring any additional representational layer or external control system.

So far, our attention has focused on the inputs received by the sensory system. However, the definition of the senses—and therefore of the sensory system of an individual—is inseparable from the purpose of the individual and the way in which that purpose is achieved. The goal of an individual is survival. Within EGO, survival is defined as compliance with the rules of the individual's organization. These rules consist of a set of $m$ categories, where each category $j$ must contain a prescribed number $k_j$ of instances. This organizational constraint defines the purpose of a particular individual and depends both on the sensory system through which the individual interacts with its environment and on the organizational rules that have been assigned to it. Once again, the universality of the mechanism becomes apparent. Not only may the $m$ categories and the corresponding values of $k_j$ differ from one individual to another, but the sensory systems themselves may also vary. Up to this point, we have described an individual as a computational counterpart of a biological organism. However, there is no reason to restrict the model to biological domains. For example, the sensory system of an individual could consist of sensors monitoring the state of an electrical power grid, or of devices tracking financial flows for trading operations. More generally, the function of an individual, as perceived by an external observer, is determined by the combination of its sensory system and its organizational rules. Consider the case of an individual designed to maintain the stability of an electrical grid. In this scenario, the organizational rules specify the conditions that define a balanced state of the grid, while the sensory system is composed of measurement instruments distributed across electrical substations. Through appropriate actuators, the individual may then respond to perturbations by increasing or decreasing power generation, reducing consumption, or redistributing energy across the network. From the perspective of EGO, this application differs from a biological organism only in the nature of its sensory apparatus and organizational constraints; the underlying cognitive mechanism remains unchanged.

Within the current version of EGO, the mechanism responsible for preserving an individual's organization is divided into two components: Manipulation and Behaviour. Manipulation corresponds to the construction of an action plan, whereas Behaviour corresponds to the execution of that plan. In the present implementation, Manipulation is addressed in a simplified manner within an abstract environment, where achieving a desired result merely requires transferring elements from one category instance to another. In reality, however, categories are multisensory and their instances correspond to objects that exist in space and time, whose constituent components cannot always be isolated or manipulated directly. As a consequence, a realistic implementation of Manipulation would likely need to be decomposed into multiple sub-manipulations, each contributing to the achievement of the desired outcome. Moreover, once transformed into Behaviours, these sub-manipulations must confront the constraints of feasibility. An action plan may be logically valid while remaining physically, temporally, or operationally impossible to execute. A further challenge therefore consists in establishing the mechanisms through which Manipulations are translated into executable Behaviours.

Interestingly, the distinction introduced here between Manipulation, understood as the internal construction of an action plan, and Behaviour, corresponding to its execution, appears consistent with emerging evidence from hippocampal physiology. Studies on hippocampal replay have shown that sequences of place-cell activity are expressed not only after navigation (reverse replay), but also before movement as forward "preplay", suggesting that internally generated sequences may contribute to the planning of forthcoming trajectories before their behavioral execution (Diba & Buzsáki, 2007).

Although EGO operates at a much higher level of abstraction and does not attempt to model hippocampal circuitry, this functional correspondence suggests that separating action planning from action execution may reflect a more general organizational principle of adaptive cognitive systems. In EGO, Manipulation is not restricted to spatial navigation, but corresponds more generally to the internal generation of ordered sequences of transformations capable of achieving a desired goal. From this perspective, hippocampal preplay may represent one biological instance of a more general computational strategy for prospective action generation.

The use of K-means should likewise not be regarded as an essential component of the architecture. In the present implementation, K-means provides a simple mechanism for extracting Common Aspects from sensory inputs. However, the theoretical framework requires only the existence of a process capable of identifying stable regularities within sensory experience. Alternative clustering or representation-learning techniques could therefore be employed without affecting the fundamental principles of EGO. In addressing these issues, the concept of Manipulation introduced in the present work may itself require revision and refinement. The objective of the current implementation was not to provide a definitive realization of Manipulation, but rather to demonstrate how such a mechanism can emerge from the interaction between categorization processes and the definition of a goal. The architecture presented here should therefore be regarded as a proof of concept establishing the feasibility of the proposed approach while leaving ample room for future developments.

Future research should focus on several directions:

1. The development of multisensory individuals capable of integrating information originating from multiple sensory modalities into unified cognitive structures.
2. The development of more sophisticated mechanisms for transforming Manipulations into Behaviours, thereby bridging the gap between abstract action planning and executable actions in complex environments.
3. The investigation of alternative mechanisms for extracting Common Aspects and generating categories from sensory experience.

More generally, the present version of EGO provides evidence for the feasibility of a universal mechanism for the categorization of computationally accessible reality. The results suggest that it is possible to construct individuals whose function, from the perspective of an observer, is determined by the interaction between a sensory system and a set of organizational constraints defining their goals. These goals can in turn be pursued through Manipulation, understood as the construction of an abstract action plan that is subsequently translated into Behaviours. From this perspective, EGO should not be viewed as an alternative to contemporary generative AI systems, but rather as an attempt to address dimensions of cognition that remain largely unexplored within current language-model-based architectures, including intrinsic grounding, operational closure, homeostatic self-maintenance, and the distinction between organization and structure.

More fundamentally, the present work suggests that cognition may be understood not as the manipulation of representations, but as the continuous preservation of an organization through structurally adaptive processes. If this hypothesis is correct, intelligence is not primarily a property of computation itself, but of the organizational principles governing computation. Naturally, the work presented here represents only an initial step toward a different conception of Artificial General Intelligence. Its significance lies less in the specific implementation of the proposed mechanisms than in demonstrating the feasibility of an approach in which cognition, goal-directed behaviour, and adaptive action emerge from recursive processes that preserve organizational identity through structural change. In this perspective, intelligence is not defined by the ability to produce correct outputs, but by the capacity of a system to maintain and regenerate its organization while interacting with its environment.

## Appendix A. The E-language structure

**E-language definition**

The four bold symbols '**0**', '**,**', '**(**', and "**)**" - which can be read as "empty", "comma", "opening bracket" and "closing bracket" - constitute the "alphabet" of E-language. They are used in bold to differentiate them from the analogue symbols of the metalanguage used to describe E-language. We call E-expression a finite sequence of such symbols. Italic uppercase letters with eventual subscripts are used as variables that represent E-expressions.

**Definition 1**. [WFFs Definition]

We call well-formed formulas [WFFs] the following E-expression:

a) **0**.
b) If $X_1, ..., X_n$ with $n \geq 1$ are WFFs, then $\mathbf{(}X_1\mathbf{,} ...\mathbf{,} X_n\mathbf{)}$ is a WFF.
c) An E-expression is a WFF if and only if it necessarily results from conditions [a] and [b].

**Example 1**.

Examples of WFFs are: **(0, 0)**; **(0)**; **(0, (0))**; **((0, (0)))** …. etc.

**Interpretation of WFFs as sets**

WFFs can be interpreted as sets built from the primitive term **0**.

**Definition 2**. [Definition of $X = Y$]

A WFF $X$ is 'equal' to a WFF $Y$ [written $X = Y$] iff one of the following conditions holds:

a) $X$ is **0** and $Y$ is **0**.
b) $X$ is $\mathbf{(}X_1\mathbf{,} ...\mathbf{,} X_m\mathbf{)}$, $Y$ is $\mathbf{(}Y_1\mathbf{,} ...\mathbf{,} Y_n\mathbf{)}$, and for each $X_i$, with $1 \leq i \leq m$, there is some $Y_j$, with $1 \leq j \leq n$ such that $X_i = Y_j$, and for each $Y_j$, with $1 \leq j \leq n$, there is some $X_i$ with $1 \leq i \leq m$ such that $Y_j = X_i$.

When and only when a WFF $X$ is not equal to a WFF $Y$ we write $X \neq Y$.

**Example 2**.

$X$ = **(0, (0), (0, (0)))** = $Y$ = **(((0), 0), 0, (0), 0)**. In fact, $X_1 = Y_2 = Y_4$ since **0** = **0** by def. 2a; $X_2 = Y_3$ because **(0)** = **(0)** by def. 2b; finally, $X_3 = Y_1$, that is **(0, (0))** = **((0), 0)**. In fact, if $U$ is **(0, (0))** and $V$ is **((0), 0)**, $U_1 = V_2$ since **0** = **0** and $U_2 = V_1$ since **(0)** = **(0)**.

We shall point out that, because of definition 2, for any WFF $X$, $\mathbf{(}X\mathbf{,} \ldots\mathbf{,} X\mathbf{)} = \mathbf{(}X\mathbf{)}$.

**Definition 3.** [Membership relation]

A WFF $X$ is a 'member' of a WFF $Y$ [and we write $X \in Y$] iff $Y = (Y_1, ..., Y_n)$ and there is some $Y_j$, with $1 \leq j \leq n$, such that $X = Y_j$.

When and only when a WFF $X$ is not a member of a WFF $Y$, we write $X \notin Y$.

The **0** symbol corresponds to the empty set. In fact, per definition 1, all WFFs are built from **0,** and thus, per definition 2, we cannot have $X \in \emptyset$ for any $X$.

For the following definitions, theorems, and proofs, we will use natural language and not the notation normally used by first-order logic. This preference is meant to avoid confusion between the symbols of the logical operators that we will define as abbreviations of WFFs, and the symbols of first-order logic operators that we should have used in the metalanguage.

**Interpretation of WFFs as statements**

In addition to representing sets, WFFs represent logical sentences if we interpret them in the following way:

1. **0** is a variable that can assume the values **T** or **F**.
2. Let $X_1, ..., X_n$ be $n$ WFFs, with $n \geq 1$, where the $n$ WFFs can assume the **T** or **F** values; if each of the $X_1, ..., X_n$ WFFs has the **T** value, then $(X_1, ..., X_n)$ has the **F** value. Otherwise, $(X_1, ..., X_n)$ has the **T** value.

By [1] and [2] a WFF can be interpreted as the NAND of propositional calculus. In fact, NAND returns the "false" [**F**] value if, and only if, all its arguments have the "true" [**T**] value. For any WFF $X$, we say that $X$ is a tautology [and we write: $\top X$] iff $X$ takes value **T** for any truth value attributed to **0**. Similarly, for any WFF $X$, we say that $X$ is a contradiction [and we write: $\bot X$] iff $X$ takes value **F** for any truth value attributed to **0**.

**Definition 4**. [Logical equivalence between WFFs]

We say that two WFFs $X$ and $Y$ E are logically equivalent when and only when $X$ and $Y$ have the same truth value, while **0** varies.

We recall that, according to Definition 2, for every WFF $X$, we have the set equality $(X, \ldots, X) = (X)$. Furthermore, $(X, \ldots, X)$ is logically equivalent to $(X)$. In fact, if the truth value of $X$ is **T**, both $(X, \ldots, X)$ and $(X)$ are **F,** and if $X$ is **F**, both $(X, \ldots, X)$ and $(X)$ are **T**. This allows us to reduce $(X, \ldots, X)$ to $(X)$. We therefore introduce the following definition.

**E-formulas**

The theory of E-language can be simplified by considering only WFFs in which there is no repetition of its members. We will call E-formulas such WFFs.

**Definition 5.** [Definition of E-formula]

Any WFF $X$ is an E-Formula iff one of these two conditions are satisfied

1. $X = \mathbf{0}$.
2. $X = \mathbf{(}X_1\mathbf{,} ...\mathbf{,} X_n\mathbf{)}$ with $X_1, ..., X_n$ which are WFFs and such as $X_i \neq X_j$ if $i \neq j$.

In other words, EFs are built on WFFs that are not equal to each other.

**Example 3**.
The WFF **(0, 0)** is not an EF, while **(0)** is an EF, even if **(0, 0)** and **(0)** are equal in terms of set and are logically equivalent as statements.

**Theorem 1.** [Equality between EFs as a 1-1 correspondence between their members]
For any EFs $\mathbf{(}X_1\mathbf{,} ...\mathbf{,} X_m\mathbf{)}$ and $\mathbf{(}Y_1\mathbf{,} ...\mathbf{,} Y_n\mathbf{)}$, $\mathbf{(}X_1\mathbf{,} ...\mathbf{,} X_m\mathbf{)} = \mathbf{(}Y_1\mathbf{,} ...\mathbf{,} Y_n\mathbf{)}$ iff, for each $X_i \in \mathbf{(}X_1\mathbf{,} ...\mathbf{,} X_n\mathbf{)}$ there is one and only one $Y_j \in \mathbf{(}Y_1\mathbf{,} ...\mathbf{,} Y_m\mathbf{)}$ such that $X_i = Y_j$, and vice versa.

**Proof.**

1. Hypothesis: $\mathbf{(}X_1\mathbf{,} ...\mathbf{,} X_m\mathbf{)}$ and $\mathbf{(}Y_1\mathbf{,} ...\mathbf{,} Y_n\mathbf{)}$ are EFs, and $\mathbf{(}X_1\mathbf{,} ...\mathbf{,} X_m\mathbf{)} = \mathbf{(}Y_1\mathbf{,} ...\mathbf{,} Y_n\mathbf{)}$.
If there is not a 1-1 correspondence indicated in the theorem, for some $X_i \in \mathbf{(}X_1\mathbf{,} ...\mathbf{,} X_m\mathbf{)}$, one $Y_j \in \mathbf{(}Y_1\mathbf{,} ...\mathbf{,} Y_n\mathbf{)}$, and one $Y_k \in \mathbf{(}Y_1\mathbf{,} ...\mathbf{,} Y_n\mathbf{)}$, should be $X_i = Y_j = Y_k$ with $j \neq k$. Therefore, by Def. 5, $\mathbf{(}Y_1\mathbf{,} ...\mathbf{,} Y_n\mathbf{)}$ would not be an EF, contradicting the hypothesis. Similarly, if $Y_j = X_i = X_k$.
2. Hypothesis: $\mathbf{(}X_1\mathbf{,} ...\mathbf{,} X_m\mathbf{)}$ and $\mathbf{(}Y_1\mathbf{,} ...\mathbf{,} Y_n\mathbf{)}$ are EFs, and for each $X_i \in \mathbf{(}X_1\mathbf{,} ...\mathbf{,} X_m\mathbf{)}$ there is one and only one $Y_j \in \mathbf{(}Y_1\mathbf{,} ...\mathbf{,} Y_n\mathbf{)}$, such that $X_i = Y_j$ and vice versa.
If it were $\mathbf{(}X_1\mathbf{,} ...\mathbf{,} X_m\mathbf{)} \neq \mathbf{(}Y_1\mathbf{,} ...\mathbf{,} Y_n\mathbf{)}$, by definition 2b for some $X_i \in \mathbf{(}X_1\mathbf{,} ...\mathbf{,} X_m\mathbf{)}$, it must be $X_i \neq Y_j$ for each $Y_j \in \mathbf{(}Y_1\mathbf{,} ...\mathbf{,} Y_n\mathbf{)}$.

Theorem 1 implies that $m = n$ is a necessary condition for $\mathbf{(}X_1\mathbf{,} ...\mathbf{,} X_m\mathbf{)} = \mathbf{(}Y_1\mathbf{,} ...\mathbf{,} Y_n\mathbf{)}$ .

By definition 2, $\mathbf{(}X\mathbf{,} Y_1\mathbf{,} ...\mathbf{,} Y_n\mathbf{)} = \mathbf{(}X\mathbf{,} \ldots\mathbf{,} X\mathbf{,} Y_1\mathbf{,} ...\mathbf{,} Y_n\mathbf{)}$. Therefore, the EF $\mathbf{(}X\mathbf{,} Y_1\mathbf{,} ...\mathbf{,} Y_n\mathbf{)}$ can be considered representative of all WFFs $\mathbf{(}X\mathbf{,} \ldots\mathbf{,} X\mathbf{,} Y_1\mathbf{,} ...\mathbf{,} Y_n\mathbf{)}$, regardless of the number of occurrences of $X$ in the WFF. Furthermore, the interpretation of any of these WFFs as a logical statement always generates the same truth value as the EF that represents them. That is, the propositional calculus is invariant within the equivalence class represented by the EF $\mathbf{(}X\mathbf{,} Y_1\mathbf{,} ...\mathbf{,} Y_n\mathbf{)}$. Therefore, all the theorems proved on EFs are valid for WFFs and therefore from this moment on we will focus only on EFs.

We can introduce the following EFs abbreviations.

**Definition 6.** [Introduction of propositional calculus operators]

a) For every WFF $X$:
   $\sim X$ stands for $\mathbf{(}X\mathbf{)}$.
b) For every tuple of WFFs $X_1, ..., X_n$, with $n \geq 1$:
   $\mathbf{(}X_1 \wedge \ldots \wedge X_n\mathbf{)}$ stands for $\sim\mathbf{(}X_1\mathbf{,} ...\mathbf{,} X_n\mathbf{)}$ and thus, for $\mathbf{((}X_1\mathbf{,} ...\mathbf{,} X_n\mathbf{))}$.
   $\mathbf{(}X_1 \vee \ldots \vee X_n\mathbf{)}$ stands for $\mathbf{(}\sim X_1\mathbf{,} ...\mathbf{,} \sim X_n\mathbf{)}$ and thus, for $\mathbf{((}X_1\mathbf{),} ...\mathbf{,} \mathbf{(}X_n\mathbf{))}$.
c) For every pair of WFFs $X_1$ and $X_2$:

$\mathbf{(}X_1 \supset X_2\mathbf{)}$ stands for $\mathbf{(}{\sim}X_1 \lor X_2\mathbf{))}$ and thus, for $\mathbf{(}X_1\mathbf{,}\ \mathbf{(}X_2\mathbf{))}$.
$\mathbf{(}X_1 \equiv X_2\mathbf{)}$ stands for $\mathbf{((}X_1\mathbf{,}\ \mathbf{(}X_2\mathbf{))} \land \mathbf{(}X_2\mathbf{,}\ \mathbf{(}X_1\mathbf{)))}$ and thus, for $\mathbf{(((}X_1\mathbf{,}\ \mathbf{(}X_2\mathbf{)),}\ \mathbf{(}X_2\mathbf{,}\ \mathbf{(}X_1\mathbf{))))}$.

It is easy to verify that the characters ~, ∧, ∨, ⊃, and ≡ can be interpreted respectively as the operators of negation, conjunction, disjunction, implication and equivalence in propositional calculus. In fact, when interpreted in this way, the abbreviations have the same truth value as abbreviated EFs.

### E-language self-referentiality

For each pair of EFs $X$ and $Y$, it is always possible to build an EF called the "Equality Evaluator" [EE], which has the following property: it is a tautology if and only if $X = Y$, it is a contradiction if and only if $X \neq Y$. Definitions 7-11 formalize the process of constructing EE.

**Definition 7.** [Definition of *Cont*[$X$]]
For every EF $X$, *Cont*[$X$] is an EF - which we call "container" of $X$ - defined in the following way:

a) *Cont*[$X$] = $X$ iff $X$ holds at least one member. That is, *Cont*[$\mathbf{(}X_1\mathbf{,}\ \ldots\mathbf{,}\ X_n\mathbf{)}$] = $\mathbf{(}X_1\mathbf{,}\ \ldots\mathbf{,}\ X_n\mathbf{)}$, with $n \geq 1$.
b) *Cont*[$X$] = $\mathbf{(}X\mathbf{)}$ iff $X$ does not hold members. That is, *Cont*[$\mathbf{0}$] = $\mathbf{(0)}$.

It is very important to note that *Cont*[$\mathbf{(0)}$] = $\mathbf{(0)}$ by def. 7a, and *Cont*[$\mathbf{0}$] = $\mathbf{(0)}$ by def. 7b. In other words, *Cont*[$\mathbf{(0)}$] = *Cont*[$\mathbf{0}$] = $\mathbf{(0)}$. *As we will see in example 4 after definition 11, this is the formal core of the self-reference theorem.*

**Definition 8.** [Definition of F[$X$, $Y$]]
F[$X$, $Y$] represents any surjective application from *Cont*[$X$] to *Cont*[$Y$].

Therefore, F[$X$, $Y$] constitutes the pairing of each member of *Cont*[$X$] with one and only one member of *Cont*[$Y$], such that each member of *Cont*[$Y$] is paired with at least one member from *Cont*[$X$]. It is very important to note that according to definitions 7 and 8, when $X$ has members and $Y$ has none, F[$X$, $Y$] pairs each member of $X$ to the symbol $\mathbf{0}$. Definitions 7 and 8 have been introduced as they allow the members of an EF $X$ to be connected with those of an EF $Y$ even when $Y$ has no members, including the possibility for $\mathbf{(0)}$ to be paired with $\mathbf{0}$. This pairing is the core of theorem 2 [the self-reference theorem], since as a logical sentence, $\mathbf{(0)}$ is the negation of $\mathbf{0}$. The system of definitions that we are illustrating in this section will allow the members of $X$ and $Y$ to be recursively connected through logical equivalence so that the contradiction $\mathbf{(0 \equiv (0))}$ is generated iff $X \neq Y$.

Obviously, there may be no surjective application from *Cont*[$X$] to *Cont*[$Y$] when the number of members of $X$ is less than those of $Y$. In this case, to maintain a pairing between the members, we must introduce a new definition.

**Definition 9.** [Definition of $\mathfrak{F}$[$X$, $Y$]]

$\mathfrak{F}[X, Y]$ stands for $F[X, Y]$ if there is a surjective application from $Cont[X]$ to $Cont[Y]$. Otherwise, it stands for $F[Y, X]$, always ordering the members in the pairs of $F[Y, X]$ having the member of $Cont[X]$ as the first member of the pair.

By definition 9, each ordered pair $<W_i, Z_j>$ of $\mathfrak{F}[X, Y]$ is such that $W_i$ belongs to $Cont[X]$ and $Z_j$ belongs to $Cont[Y]$, even when there are no surjective applications from $Cont[X]$ to $Cont[Y]$, but only from $Cont[Y]$ to $Cont[X]$.

Let us now define $\mathbf{(}X = Y\mathbf{)}$, which is the EF that will be the subject of the self-reference theorem.

**Definition 10.** [Definition of $\mathfrak{F}$^$[X = Y]$]
For every $X$ and $Y$ EFs:

a) If $Cont[X] = \mathbf{(}U\mathbf{)}$ and $Cont[Y] = \mathbf{(}V\mathbf{)}$:
$\mathfrak{F}$^$[X = Y]$ stands for $\mathbf{(}U = V\mathbf{)}$
where $<U, V>$ is the only pair of $\mathfrak{F}[X, Y]$.

b) Otherwise:
$\mathfrak{F}$^$[X = Y]$ stands for $\mathbf{((}W_1 = Z_1\mathbf{)} \wedge \ldots \wedge \mathbf{(}W_k = Z_k\mathbf{))}$
where $<W_1, Z_1>, \ldots, <W_k, Z_k>$ are the $k$ pairs of a $\mathfrak{F}[X, Y]$..

**Definition 11.** [Definition of $\mathbf{(}X = Y\mathbf{)}$]
For any EFs $X$ and $Y$:

a) If $Cont[X] = Cont[Y] = \mathbf{(0)}$, then:
$\mathbf{(}X = Y\mathbf{)}$ stands for $\mathbf{(}X \equiv Y\mathbf{)}$.

b) If $Cont[X] = \mathbf{(0)} \neq Cont[Y]$ or $Cont[X] \neq \mathbf{(0)} = Cont[Y]$ then:
$\mathbf{(}X = Y\mathbf{)}$ stands for $\mathfrak{F}$^$[X = Y]$

c) If $Cont[X] \neq \mathbf{(0)}$ and $Cont[Y] \neq \mathbf{(0)}$, then:
$\mathbf{(}X = Y\mathbf{)}$ stands for $\mathbf{(}\mathfrak{F}_1$^$[X = Y] \vee \ldots \vee \mathfrak{F}_n$^$[X = Y]\mathbf{)}$
where $\mathfrak{F}_1[X, Y], \ldots, \mathfrak{F}_n[X, Y]$ are the $n$ $\mathfrak{F}[X, Y]$ existing between $Cont[X]$ and $Cont[Y]$.

**Example 4.**
$X = \mathbf{0}$ and $Y = \mathbf{(0)}$. Then $Cont[X] = \mathbf{(0)}$ by def. 7b, and $Cont[Y] = \mathbf{(0)}$ by def. 7a. Then $\mathbf{(}X = Y\mathbf{)}$ stands for $\mathbf{(0 \equiv (0))}$ by def. 11a, where $\mathbf{(0 \equiv (0))}$ *is a contradiction. All that follows is to show that an E-formula X is set-theoretically not equal to an E-formula Y if and only the E-formula that evaluates the equality between X and Y is a logical conjunction of which at least one of the conjuncts is the contradiction* $\mathbf{(0 \equiv (0))}$.

**Example 5.**
Stand $X = \mathbf{(((0)), 0)}$ and $Y = \mathbf{(0, (0))}$. Then, $\mathbf{((((0)), 0) = (0, (0)))}$

stands for: $\mathbf{(((((0)) = 0) \wedge (0 = (0))) \vee ((((0)) = (0)) \wedge (0 = 0)))}$ def. 11c.

stands for: $\mathbf{((((0) = 0) \wedge (0 \equiv (0))) \vee (((0) = 0) \wedge (0 \equiv 0)))}$
def. 11b and 11a.

stands for: $\mathbf{((((0) \equiv 0) \wedge (0 \equiv (0))) \vee (((0) \equiv 0) \wedge (0 \equiv 0)))}$
def. 11a.

The final EF is a disjunction of two logical conjunctions. Each conjunction is a contradiction and, therefore, the entire EF is a contradiction. The self-reference theorem, which is stated and proved later, states that this happens because $X$ and $Y$ in this case are set-theoretically unequal. If, however, they had been set-theoretically equal, at least one logical conjunction would have been a tautology and therefore the entire EF would have been a tautology. The self-reference theorem states that this happens for any pair of $X$ and $Y$. To further clarify Example 5 and hence also the logic of the self-reference theorem, we can represent $X$ and $Y$ as rooted trees as illustrated in Figure 4.

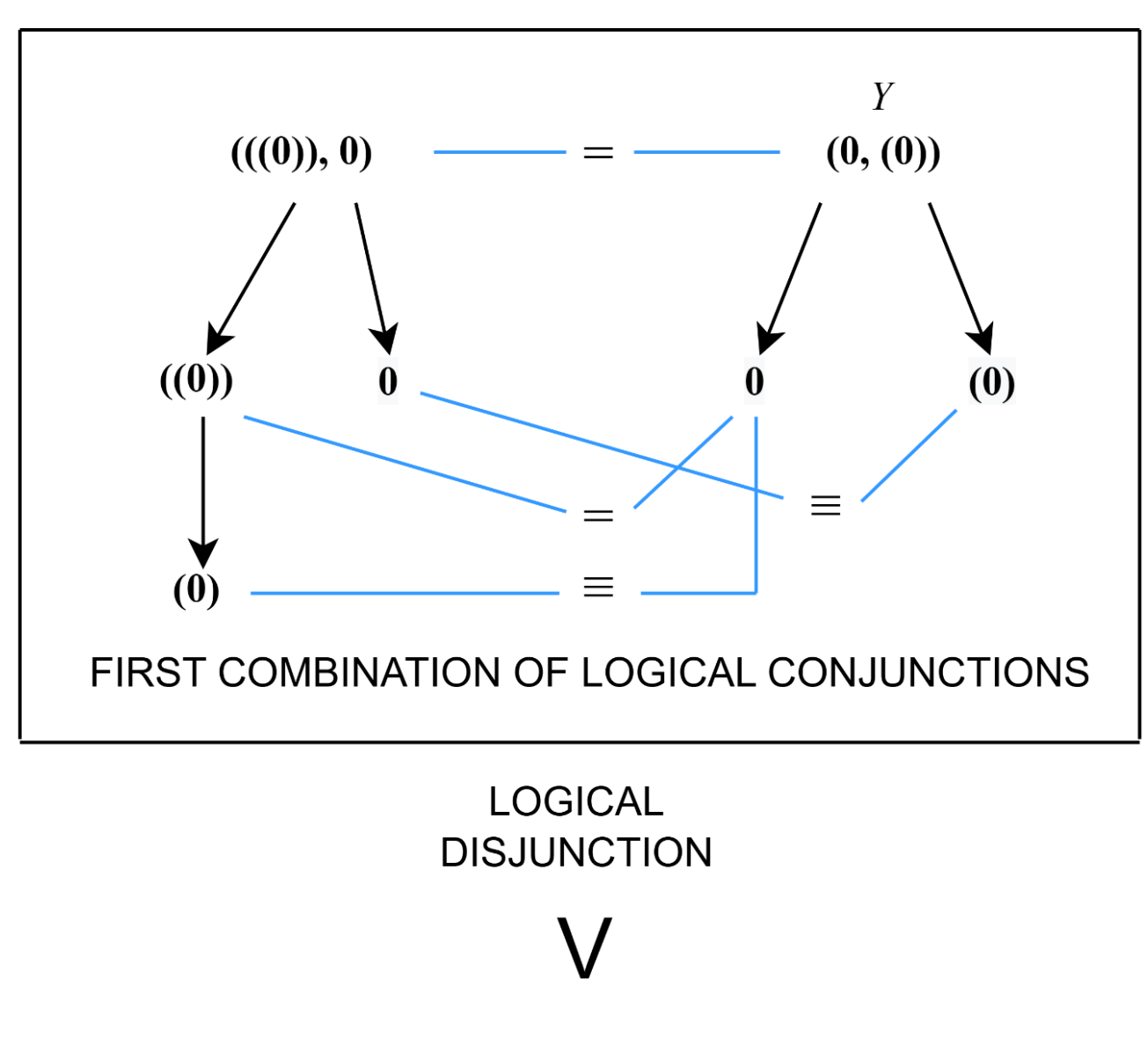

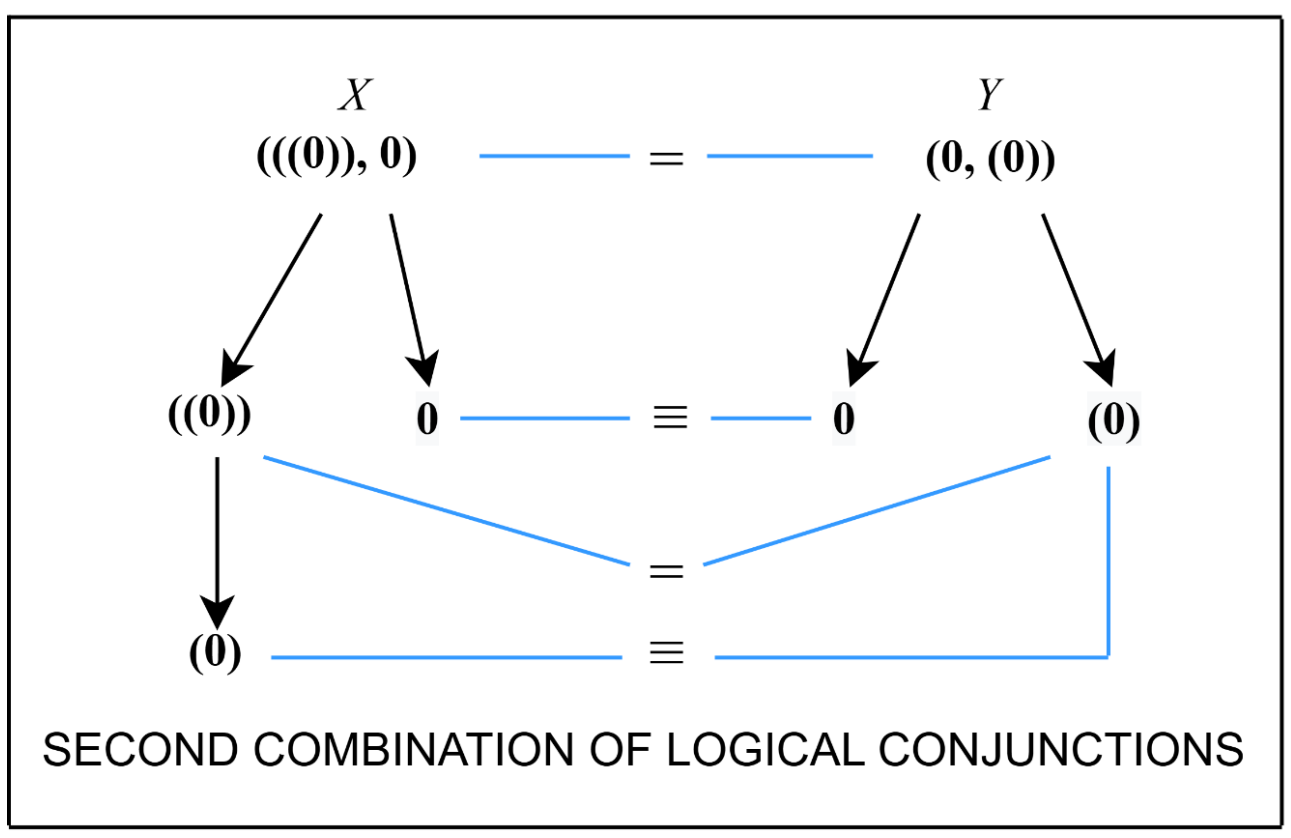

**Figure 4**. Graphical representation of example 5.

The black arrows indicate the edges of the graph. The blue lines indicate the associations, via the equality symbol, between the nodes of $X$ at a given hierarchical level with those of $Y$ at the same hierarchical level. When both $X$ and $Y$ are **0** or **(0)**, equality is replaced by equivalence. Note that in the formula of Example 5 there are two combinations of logical conjunctions, which are linked by a disjunction, as represented by the OR [V] in the Figure 4.
We now turn to the proof of a lemma of the self-reference theorem.

**Definition 12.**
Let us call M the set of EFs defined as:

a) **0** belongs to M.
b) If an EF $X$ belongs to M, then $(X)$ belongs to M.
c) An EF belongs to M if and only if it necessarily results from conditions [a] and [b].

**Lemma of Theorem 2**
Let $Y$ be an EF belonging to M. $\mathbf{0} = Y$ iff the EF $(\mathbf{0} = Y)$ is a tautology; and $\mathbf{0} \neq Y$ iff the EF $(\mathbf{0} = Y)$ is a contradiction. That is: $\mathbf{0} = Y$ iff $\top (\mathbf{0} = Y)$; and $\mathbf{0} \neq Y$ iff $\bot (\mathbf{0} = Y)$.

**Proof**
By induction on M definition.
Hypothesis: $\mathbf{0} = Y$.

1. $Cont[Y] = Cont[\mathbf{0}] = (\mathbf{0})$. — hypothesis and def. 7b.
2. $(\mathbf{0} = Y)$ stands for $(\mathbf{0} \equiv \mathbf{0})$. — 1 and def. 11a.
3. $\top (\mathbf{0} \equiv \mathbf{0})$. — Def. 6.
4. $\top (\mathbf{0} = Y)$. — 2 and 3.
5. $\mathbf{0} = Y$ implies $\top (\mathbf{0} = Y)$. — hypothesis and 4.

Hypothesis: $\mathbf{0} \neq Y = (\mathbf{0})$.

1. $Cont[Y] = Cont[(\mathbf{0})] = (\mathbf{0})$. — hypothesis and def. 7a.
2. $Cont[\mathbf{0}] = (\mathbf{0})$. — def. 7b.
3. $(\mathbf{0} = Y)$ stands for $(\mathbf{0} \equiv (\mathbf{0}))$. — 1, 2, and def. 11a.
4. $\bot (\mathbf{0} \equiv (\mathbf{0}))$. — Def. 6.
5. $\bot (\mathbf{0} = Y)$. — 3 and 4.
6. $\mathbf{0} \neq Y = (\mathbf{0})$ implies $\bot (\mathbf{0} = Y)$. — hypothesis and 5.

Hypothesis: $\mathbf{0} \neq Y = (Y_1)$ and $Y_1 \neq \mathbf{0}$.

1. $Cont[Y] = Cont[(Y_1)] = (Y_1) \neq (\mathbf{0})$. — hypothesis and def.7a.
2. $Cont[\mathbf{0}] = (\mathbf{0})$. — def. 7b.
3. $(\mathbf{0} = Y)$ stands for $\mathfrak{F}^\wedge[\mathbf{0} = (Y_1)]$. — 1, 2, and def. 11b.
4. $(\mathbf{0} = Y)$ stands for $(\mathbf{0} = Y_1)$. — 1,2, 3, and def. 10a.
5. $\mathbf{0} \neq Y_1$. — hypothesis.
6. $\bot (\mathbf{0} = Y_1)$. — inductive hypothesis.
7. $\bot (\mathbf{0} = Y)$. — 4 and 6.
8. $\mathbf{0} \neq Y = (Y_1)$ and $Y_1 \neq \mathbf{0}$ implies $\bot (\mathbf{0} = Y)$. — hypothesis and 7.

We can finally state and prove the self-reference theorem.

**Theorem 2.** [Self-reference theorem]
Given any two $X$ and $Y$ EFs, $X = Y$ iff $(X = Y)$ is a tautology; $X \neq Y$ iff $(X = Y)$ is a contradiction. That is: $X = Y$ iff $\top (X = Y)$; $X \neq Y$ iff $\bot (\mathbf{0} = Y)$.

**Proof**
Let us proceed by induction on WFF definition [def. 1].
If $X = \mathbf{0}$ and $Y$ is an EF of M the theorem is true for the lemma.
Hypothesis: $\mathbf{0} = X \neq Y = (Y_1, \ldots, Y_l)$ where $l \geq 2$.

1. $Cont[X] = Cont[\mathbf{0}] = (\mathbf{0})$. — hyp. and def. 7b.
2. $Cont[Y] = Cont[(Y_1, \ldots, Y_l)] = (Y_1, \ldots, Y_l)$. — hyp. and def. 7a.
3. $(X = Y)$ stands for $\mathfrak{F}^{\wedge}[\mathbf{0} = Y]$. — def. 11b.
4. $\mathfrak{F}^{\wedge}[\mathbf{0} = Y]$ stands for $((\mathbf{0} = Y_1) \wedge \cdots \wedge (\mathbf{0} = Y_l))$. — 1, 2, and def. 10b.
5. $Y_i \neq Y_j$, where $i \neq j$. — Def. 5.
6. $\mathbf{0} \neq Y_i$ or $\mathbf{0} \neq Y_j$. — 5.
7. $\bot (\mathbf{0} = Y_i)$ or $\bot (\mathbf{0} = Y_j)$. — 6 and inductive hyp.
8. $\bot ((\mathbf{0} = Y_1) \wedge \cdots \wedge (\mathbf{0} = Y_l))$. — 7 and Def. 6.
9. $\bot \mathfrak{F}^{\wedge}[\mathbf{0} = Y]$. — 4 and 8.
10. $\bot (X = Y)$. — 3 and 9.
11. $\mathbf{0} = X \neq Y = (Y_1, \ldots, Y_l)$ implies $\bot (X = Y)$. — hypothesis and 10.

Hypothesis: $X = (X_1, \ldots, X_m) = Y = (Y_1, \ldots, Y_n)$.

1. $Cont[X] = (X_1, \ldots, X_m) = Cont[Y] = (Y_1, \ldots, Y_n)$ — hyp. and def. 7a.
2. $Cont[X] = (U_1, \ldots, U_k) = Cont[Y] = (V_1, \ldots, V_k)$ and $U_1 = V_1, \ldots, U_k = V_k$. — 1 and theorem 1.
3. $\top \mathfrak{F}_i{}^{\wedge}[X = Y]$ for some $\mathfrak{F}_1[X, Y], \ldots, \mathfrak{F}_n[X, Y]$. — 2, def. 10b, ind. hyp.
4. $\top (\mathfrak{F}_1{}^{\wedge}[X = Y] \vee \ldots \vee \mathfrak{F}_n{}^{\wedge}[X = Y])$. — 3 and Def. 6.
5. $(X = Y)$ stands for $(\mathfrak{F}_1{}^{\wedge}[X = Y] \vee \ldots \vee \mathfrak{F}_n{}^{\wedge}[X = Y])$. — def. 11c
6. $\top (X = Y)$. — 4 and 5.
7. $X = (X_1, \ldots, X_m) = Y = (Y_1, \ldots, Y_m)$ implies $\top (X = Y)$. — hypothesis and 6.

Hypothesis: $X = (X_1, \ldots, X_m)$, $Y = (Y_1, \ldots, Y_l)$, and $X \neq Y$.

1. $Cont[X] = (X_1, \ldots, X_m) \neq Cont[Y] = (Y_1, \ldots, Y_l)$. — hyp. and def. 7a.
2. For every $\mathfrak{F}_1[X, Y], \ldots, \mathfrak{F}_n[X, Y]$, there is a $X_i$ and a $Y_j$ such that $X_i \neq Y_j$ — 1, def. 2, def. 9.
3. $\bot \mathfrak{F}_1{}^{\wedge}[X = Y], \ldots, \bot \mathfrak{F}_n{}^{\wedge}[X = Y]$. — 2 and inductive hyp.
4. $\bot (\mathfrak{F}_1{}^{\wedge}[X = Y] \vee \ldots \vee \mathfrak{F}_n{}^{\wedge}[X = Y]))$. — 3 and Def. 6.
5. $(X = Y)$ stands for $(\mathfrak{F}_1{}^{\wedge}[X = Y] \vee \ldots \vee \mathfrak{F}_n{}^{\wedge}[X = Y]))$. — 1 and def. 11c.
6. $\bot (X = Y)$. — 4 and 5.
7. $X = (X_1, \ldots, X_m)$, $Y = (Y_1, \ldots, Y_l)$, and $X \neq Y$ implies $\bot (X = Y)$.hypothesis and 6.

**Example 6**.

In Example 5, we have seen that $((0) = (0, (0)))$ stands for $((((0) \equiv 0) \wedge ((0) \equiv 0)) \vee ((0 \equiv 0) \wedge (0 \equiv (0)))$. Given that $((((0) \equiv 0) \wedge ((0) \equiv 0)) \vee ((0 \equiv 0) \wedge (0 \equiv (0)))$ is a contradiction, we have the set inequality $(0) \neq (0, (0))$ and the contradiction $\perp ((0) = (0, (0)))$.

**EFs as set-theoretic relations and operations**

Theorem 2 reduces the equality relation between sets to a propositional calculus. We shall call EF $(X = Y)$ an "equality evaluator" between $X$ and $Y$. The equality evaluator is an EF that states whether EF $X$, interpreted as a set, is equal to EF $Y$ interpreted as a set. Starting from the equality evaluator, it is easy to define the evaluators of all set relations and operations.

Let us show, for example, the EFs that express the relations of membership and inclusion, and then those that express the operations of intersection and union.

**Definition 13.** [Definition of $(X \in Y)$]

a) For every EF $X$ and every EF $Y$ such that $Y = (Y_1)$:
$(X \in Y)$ stands for $(X = Y_1)$;

b) For every EF $X$ and every EF $Y$ such that $Y = (Y_1, \ldots, Y_n)$ where $n > 1$,
$(X \in Y)$ stands for $((X = Y_1) \vee \ldots \vee (X = Y_n))$.

By theorem 2, the EF represented by $(X \in Y)$ is a tautology when set $X$ is a member of set $Y$, and it is a contradiction when set $X$ is not a member of set $Y$. In fact, by def. 3, if $X$ is a member of $Y$, at least one of the equalities $X = Y_1, \ldots, X = Y_n$, must be true. For theorem 2, at least one of the EFs $(X = Y_1), \ldots, (X = Y_n)$ must be a tautology. By Def. 6, $((X = Y_1) \vee \ldots \vee (X = Y_n))$ is a tautology. If, instead, $X$ is not a member of $Y$, all inequalities $X \neq Y_1, \ldots, X \neq Y_n$ must be true. Therefore, for theorem 2, all EFs $(X = Y_1), \ldots, (X = Y_n)$ must be contradictions. Thus, for Def. 6, $((X = Y_1) \vee \ldots \vee (X = Y_n))$ is a contradiction.

**Definition 14.** [Definition of $(X \subseteq Y)$]

For every EF $X = (X_1, \ldots, X_m)$ and EF $Y = (Y_1, \ldots, Y_n)$, with $n \geq 1$ and $m \geq 1$:

$(X \subseteq Y)$ stands for $((X_1 \in Y) \wedge \ldots \wedge (X_m \in Y))$.

According to def. 14, $(X \subseteq Y)$ represents an EF only when $X \neq \mathbf{0}$ and $Y \neq \mathbf{0}$. By theorem 2, Def. 6, and def. 13, the EF represented by $(X \subseteq Y)$ is a tautology when $X$ is a *subset* of $Y$, and it is a contradiction when $X$ is not a *subset* of $Y$.

**Definition 15.** [Definition of $(X \cap Y)$]

For every $X = (X_1, \ldots, X_m)$ and $Y = (Y_1, \ldots, Y_n)$, with $m \geq 1$ and $n \geq 1$, we indicate with $(X \cap Y)$ the EF making the following EF tautological:

$$(((X_1 \in (X \cap Y)) \equiv (X_1 \in Y)) \wedge \ldots \wedge ((X_m \in (X \cap Y)) \equiv (X_m \in Y))).$$

As per theorem 2, Def. 6, and def. 13, EF **(**$X \cap Y$**)** represents the set *intersection* between $X$ and $Y$.

**Definition 16.** [Definition of **(**$X \cup Y$**)**]

For every $X$ = **(**$X_1$**, ...,** $X_m$**)** and $Y$ = **(**$Y_1$**, ...,** $Y_n$**)**, with $m \geq 1$ and $n \geq 1$, we indicate with **(**$X \cup Y$**)** the EF making the following EF tautological:

$$((X_1 \in (X \cup Y)) \wedge \ldots \wedge (X_m \in (X \cup Y)) \wedge (Y_1 \in (X \cup Y)) \wedge \ldots \wedge (Y_n \in (X \cup Y))).$$

As per Theorem 2, Def. 6, and def. 13, EF **(**$X \cup Y$**)** expresses the *union* between sets $X$ and $Y$.

## Appendix B. Conversion of a bit sequence to an E-formula

An E-formula represents a binary number if it begins with the **(((0)), 0)** E-formula. This E-formula is named "Radix". We codify the digit 0 and 1 as following:

**((((0)), 0), 0)** = 0

**((((0)), 0), (0))** = 1

We use the number of brackets surrounding the 0 and 1 to represent the position of 0 and 1 in a binary number. For example, in the number 100, the **((((((0)), 0), (0))))** E-formula represents the digit 1 in the third position of the binary string. In the same way, the digit 0 in second and first positions are respectively the **(((((0)), 0), 0))** and **((((0)), 0), 0)** E-formulas. So the number 100 is represented by the **((((((0)), 0), (0)))), (((((0)), 0), 0)), ((((0)), 0), (0)))** E-formula. We call "E-binary" an E-formula that encodes a binary number in the above way.

We now show the algorithm used to extract the decimal representation of an E-binary, writing snippets of java code. Let's now define a loop value variable as an int, and let's give this variable the value 0 at the beginning of each calculation:

```
int loopValue = 0;
```

Let's now define another variable, loopNumber such that:

```
int loopNumber = 0 if ((((0)), 0), 0)
int loopNumber = 1 if ((((0)), 0), (0))
```

Now let's define the internal loop:

```
for(int i = 0; i <= loopNumber; ++){
          loopValue = loopValue + i;}
```

It's easy to show that if `loopNumber = 0`, the `loopValue` at the end of the loop will be the same as at the beginning. In fact, the loop will be repeated just once, and as at the first cycle is i = 0, it follows that:

```
loopValue =  loopValue + 0
```

On the other hand, if `loopNumber = 1`, the loop will be repeated two times, and the second time will be `i = 1`, so that the `loopValue` variable will be increased of one unit:

```
loopValue =  loopValue + 0  -> first cycle
loopValue =  loopValue + 1  -> second cycle
```

Now, according to Zermelo's theory of ordering (1967: 202-204), let's interpret the brackets around the expression **((((0)), 0), 0)** and **((((0)), 0), (0))** as the number of external loops each of those so defined:

```
for(int i = 0; i <= loopNumber; ++){
     …. .
}
```

where the internal instruction of the loop can be either another external loop, or the internal loop when we arrive at the first red bracket. For example **((((((0)), 0), (0))))** becomes:

```
for(int i = 0; i <= loopNumber; ++){
      for(int i = 0; i <= loopNumber; ++){
            for(int i = 0; i <= loopNumber; ++){
                  loopValue = loopValue + i;
            }
      }
}
```

Now, let's take for example the number 101. This binary number will be translated in E-binary adding external brackets to 0 and 1 every time we move from the right digits of the number to the left, starting with zero brackets for the first digit:

**(((((((0)), 0), (0)))), (((((0)) ,0), 0)), ((((0)), 0), (0)))**

The first loop starts with these parameters:

```
int loopValue = 0;
int loopNumber = 1;
```

and the loop it defines is:

```
for(int i = 0; i <= loopNumber; ++){
      for(int i = 0; i <= loopNumber; ++){
            for(int i = 0; i <= loopNumber; ++){
                  loopValue = loopValue + i;
            }
      }
}
```

The `loopValue` at the end of the loop is `loopValue = 4`.

The second loop starts with these parameters:

```
int loopValue = 4;
int loopNumber = 0;
```

and the loop it defines is:

```
for(int i = 0; i <= loopNumber; ++){
      for(int i = 0; i <= loopNumber; ++){
            loopValue = loopValue + i;
      }
```

```
}
```

The `loopValue` at the end of the loop is `loopValue = 4`. The third loop starts with these parameters:

```
int loopValue = 4;
int loopNumber = 1;
```

and the loop it defines is:

```
for(int i = 0; i <= loopNumber; ++){
          loopValue = loopValue + i;
}
```

The `loopValue` at the end of the loop is `loopValue = 5`.

## Appendix C. From Common Aspects to Categories

To explain how the process of forming categories takes place starting from the common aspects we will use an example, since this will make the process clearer. Suppose you want to build a category from these common aspects

A, A, C, C, C, D, D, E

Each letter listed above represents an archetype of a common aspect. We notice that some common aspects are present more than once, in particular C is present three times, A and D are present twice, while E is present only once. The first operation performed by the algorithm is to store in a Java Map the archetypes of the common aspects present and the number of times they occur. A Map in Java is a collection of entries, each consisting of two linked parts: a key and a value. In our case, the key contains the archetype of the common aspect, and the value represents the number of times it occurs. In the example we are looking at, the Java Map will contain the following key–value pairs

| Key | E | A | D | C |
|---|---|---|---|---|
| Value | 1 | 2 | 2 | 3 |

For ease of reading in the remainder of this paragraph, the meaning of the archetype will be highlighted in gray.

**Value = 3**
We immediately notice that the common aspect that has the most occurrences is C, which appears three times. The first operation that will be done will be to create an archetype of the category that has as its meaning the name of the common aspect C. In fact, the archetype of the category that contains in its meaning the name of the common aspect C indicated as N_CA[C] will be

Cat[C] = (N_Cat[C], N_CA[C])

In turn, the name N_CA[C] refers back to the archetype whose meaning contains the E-binary of the common aspect C. We will see later how to represent the fact that Cat[C] is present three times in the meaning of the final category we want to build.

**Value = 2**

We see that A and D are both present twice. Similarly to the previous case, we can construct the archetype of the category containing the common aspect A and that of the category containing the common aspect D. The archetypes of the categories Cat[A], Cat[D]

Cat[A] = ( N_Cat[A], N_CA[C] )

Cat[D] = ( N_Cat[D], N_CA[D] )

Since A and D both appear twice, we can create a new category Cat[A,D] that contains in its meaning the names of the two categories Cat[A], Cat[D] as its elements. The general rule is to group together all the leaf categories (the ones containing a Common Aspect in its meaning) that are present the same number of times for creating a new category. This category has as elements the names of the categories that occur the same number of times. In our case, therefore, we will have

Cat[A,D] = ( N_Cat[A,D], N_Cat[A], N_Cat[D] )

Clearly Cat[A,D] contains only one element Cat[A] and only one element Cat[D]. We know instead that these elements are present twice in the meaning of the category we want to build. Again, we will explain later how to take account of this circumstance.

**Value = 1**

The common aspect E is present only once. Similarly to before, we build the category Cat[E] which contains the common aspect E. In fact, from the archetype of the common aspect E we get

Cat[E] = ( N_Cat[E], N_CA[E] )

Let's now explain the rule that returns the number of times a common aspect appears in a category. In the examples we are developing we have seen that common aspect C and the consequent category Cat[C] must appear three times in the main category we want to build. Starting from the category Cat[C], we then build the category Cat[Cat[C]] which will be defined as follows:

Cat[Cat[C]] = ( N_Cat[Cat[C]], ( (N_Cat[C]), N_Cat[C]) )

We have therefore defined a new category that contains twice, in its leaves, the common aspect C. We then continue the iteration, and generate a new category Cat[Cat[Cat[C]]] that will be defined as follows

Cat[Cat[Cat[C]]] = ( N_Cat[Cat[Cat[C]]], ((N_Cat[Cat[C]]), N_Cat[C]) )

It should be evident that the recursion rule we are using is

$EF_i[X] = (EF_{i-1}[X], Cat[X])$

where $EF_{i-1}[X]$ is the EF of category archetype at the previous step. This is a fairly intuitive rule: at each step, a category is created whose meaning is the previous iteration **nand** the category from which it originates. Representing $(EF_{i-1}[X], Cat[X])$ with a tree diagram, it results in leaves, each consisting of the E-formula Cat[X]. From the example we are analysing the common aspects A and D, present in the category Cat[A,D], appear twice in the main category. Using the rule just defined, we obtain the category Cat[Cat[A,D]], defined as follows

Cat[Cat[A,D]] = ( N_Cat[Cat[A,D]], ((N_Cat[A,D]), N_[A,D]) )

The category Cat[Cat[A,D]] contains, in its leaves, both the common aspect A and the common aspect D twice. We can therefore stop the iteration. Finally, the common aspect E appears only once in the category we want to construct; the category Cat[E] already satisfies this condition, so the previously defined rule is not applied to it. The final category Cat[A, A, C, C, C, D, D, E] will therefore be the archetype

Cat[A, A, C, C, C, D, D, E] = (N_Cat[A, A, C, C, C, D, D, E], N_Cat[E],N_Cat[Cat[A,D], N_Cat[Cat[Cat[C]]])

## Appendix D. From instances to categories

We have defined a category as a set of common aspects. The individuals construct the category representation using archetypes. In Figure 3 we have seen how each leaf of the category archetype is a common aspect archetype containing the E-binary value of its centroid in its meaning. Grouping all the values of the centroids in the leaves of the archetypus, what we find is an abstract representation of an instance of this category, having the centroid as actual value of each common aspect. Let us see the mechanism with an example using the category Cat[A, A, C, C, C, D, D, E] of the previous appendix. We use the notation CA(X) for the common aspect X and C_CA(X) for its actual centroid numerical value. Let us assume the following values

C_CA(A) = 100
C_CA(C) = 300
C_CA(D) = 400
C_CA(E) = 500

From the list defining the category Cat[A, A, C, C, C, D, D, E] we have the following list of centroids

Category Centroids = [C_CA(A), C_CA(A), C_CA(C), C_CA(C), C_CA(C), C_CA(D), C_CA(D), C_CA(E)]

Replacing each centroid symbol with its actual value we obtain the centroid values of the category expressed in decimal for readability

Category Centroid Values = [100,100,300,300,300,400,400,500]

The category centroid values shown above, once transformed in E-binary, are all present in the meaning of the common aspects of the category Cat[A, A, C, C, C, D, D, E] archetype, representing the abstract instance of the category. The phrase 'abstract instance' is used in the sense that for obtaining an actual instance of this category we have to replace the centroid values with the actual values of the instance. For example the following values

[98,102,305,304,299,402,408,504]

are an instance of the category Cat[A, A, C, C, C, D, D, E]. How does the individual know that this is an instance of the category Cat[A, A, C, C, C, D, D, E]? Because the Individual uses the Kmeans model to find the centroid of each of the [98,102,305,304,299,402,408,504] numbers. Once the centroids are known, it's possible to find all the common aspects related to those centroids. In other words, starting from the incoming stimuli (we show them already converted to decimal from their binary form because Kmeans ml in Sparkl only accepts decimal values) [98,102,305,304,299,402,408,504] the Kmeans model will returns to the individual the centroids [100,100,300,300,300,400,400,500]. Then the individual transforms the centroids in E-binaries. From these centroids, transformed in E-binaries, he finds the common aspects having the values of these centroids in their meaning [CA(A),CA(A),CA(C),CA(C),CA(C),CA(D),CA(D),CA(E)]. After that, the individual searches in its memory. If a category having common aspects [CA(A),CA(A),CA(C),CA(C),CA(C),CA(D),CA(D),CA(E)] in its leaves already exists, he only adds the materialized archetype of the instance to the structure. If not, both the category archetype and the materialized archetype of the instance are added to the structure. The category archetype is constructed according to the rules seen in the previous section.

What has been described so far is only one of the ways in which categories are generated. Individuals are fully capable of constructing new categories they have never encountered, including in the meaning of these new categories either the names of already known categories or the names of common aspects they are familiar with. Moreover, once a new category is generated, through manipulation and behaviors the individual is able to generate instances of these categories from instances of categories belonging to the emerging results (emerging results are the result of environmental shocks) present in its structure.

For example, given the Categories Cat(A), Cat(B), Cat(C), it is possible to build a new category Cat(D), having Cat(A), Cat(B) and Cat(C) as elements. It will be an archetype Cat(D) having the names of the archetypes Cat(A), Cat(B), Cat(C) in its meaning:

Cat[D] = ( N_Cat[D], (N_Cat[A], N_Cat[B], N_Cat[C]) )

Once the archetype of the category is built, it's possible to create an instance of this category by substituting the centroid values of the common aspects of its leaves, with the real values of common aspects of instances of other categories present in the structure. For example, if we want to build an instance of Cat(D), and we know that both Cat(D) and Cat(K) have the common aspect CA(N) in their leaves, we can pick the CA(N) value of an instance of Cat(K) and use it to build an instance of Cat(D).

In practice, the EGO algorithms allow individuals to generate entirely new, unseen categories and to produce instances of these categories. This makes EGO a truly generative AI. Starting from the known universe, each individual is capable of generating a new, imagined universe it has never encountered—an imagined universe that it can then instantiate.

## Appendix E. An example of Manipulation

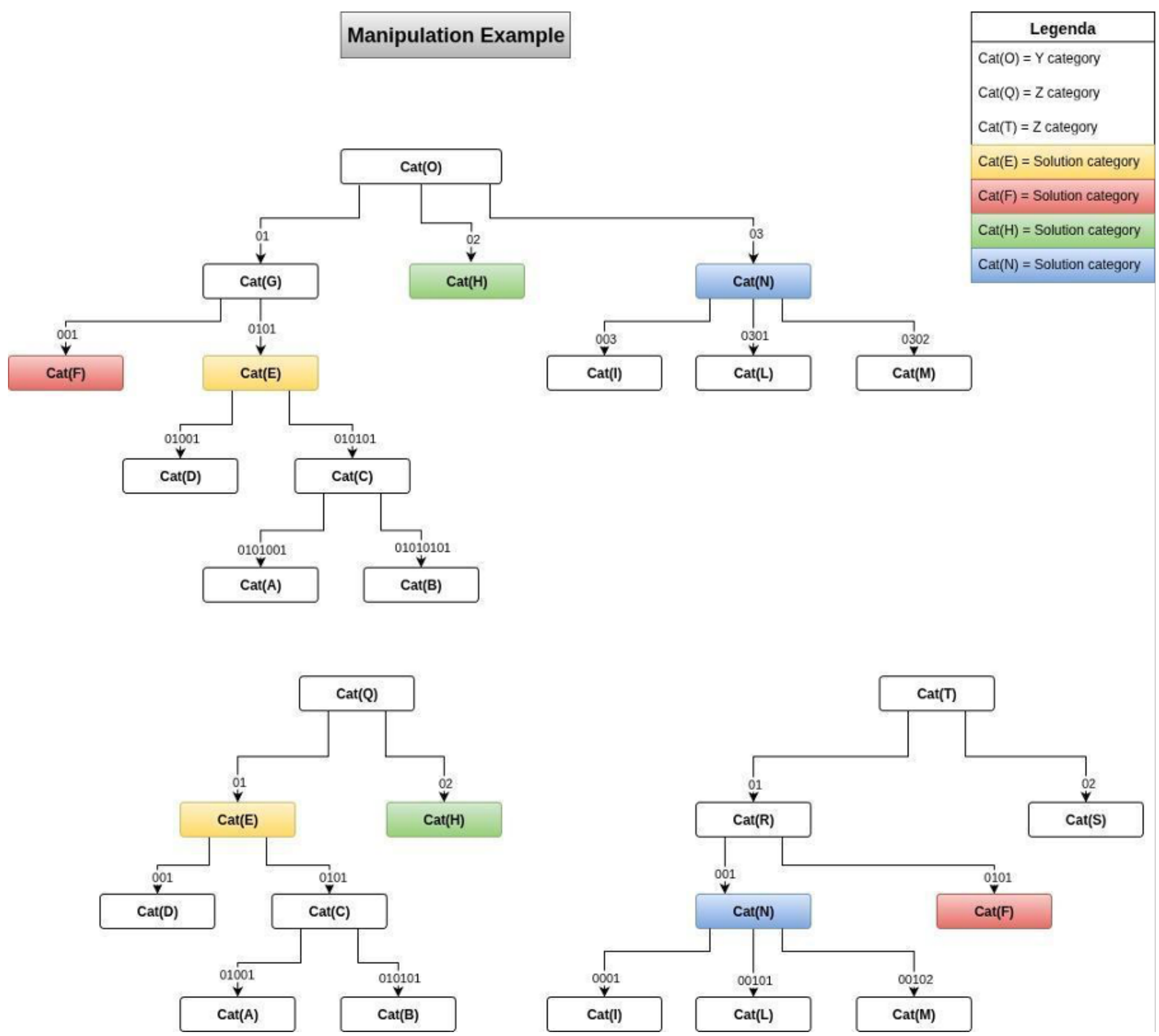

**Figure 5** An example of Manipulation.

Before presenting the example, an important point must be clarified. The Manipulation and Behaviour algorithms require a mechanism for uniquely identifying the nodes of the tree structures associated with E-formulas. The algorithm used for this purpose is described in Appendix F. In Figure 5, the node Cat(E) located at position 01 in the Cat(Q) archetype is selected and transferred to position 0101 in the Cat(O) archetype. Similarly, the node Cat(F) located at position 0101 in the Cat(T) archetype is selected and transferred to position 001 in the Cat(O) tree. Examining the Cat(O) archetype in the Figure, the newly transferred elements Cat(E) at position 0101 and Cat(F) at position 001 are both children of the node Cat(G) at position 01. Since Cat(E) and Cat(F) are the only elements of category Cat(G)—recall that every node in a category archetype tree is itself a category—the node Cat(G) can be reconstructed by placing the names (Name_Cat(E), Name_Cat(F)) in the Meaning field of the Cat(G) archetype. The node Cat(G) is thus generated. The Cat(O) archetype contains two additional elements: Cat(H) at position 02 and Cat(N) at position 03. These nodes are initially empty. They are reconstructed by transferring Cat(H) from position 02 of the Cat(Q) archetype to position 02 of the Cat(O) archetype, and Cat(N) from position 001 of the Cat(T) archetype to position 03 of the Cat(O) archetype. At this point, all three elements—Cat(G), Cat(H), and Cat(N)—are available. Consequently, category Cat(O) can be reconstructed by placing their names, (Name_Cat(G), Name_Cat(H), Name_Cat(N)), in the Meaning field of the root node of Cat(O). The category archetype Cat(O) is therefore fully reconstructed.

All operations involved in this process—including the search for matching nodes between the target category archetype Cat(O) and the source archetypes Cat(Q) and Cat(T), as well as the subsequent reconstruction steps—are recorded in a Manipulation archetype. The Manipulation archetype thus constitutes a set of instructions that can later be executed by Behaviours to construct an instance of the target category.

## Appendix F. Codification algorithm of a rooted tree

In this Appendix we present a one-one function assigning a finite decimal number at each node of a rooted tree, where each node has a finite number of children. This result is extensively used by EGO because it allows every subformula of any E-formula to be uniquely identified by one and only one decimal. A very simple example of how this result applies to E-formulas is shown in Figure 6. We present this figure also to share with the reader the imagined visualization that guided the formulation of the definitions, theorems, and their proofs found in the text.

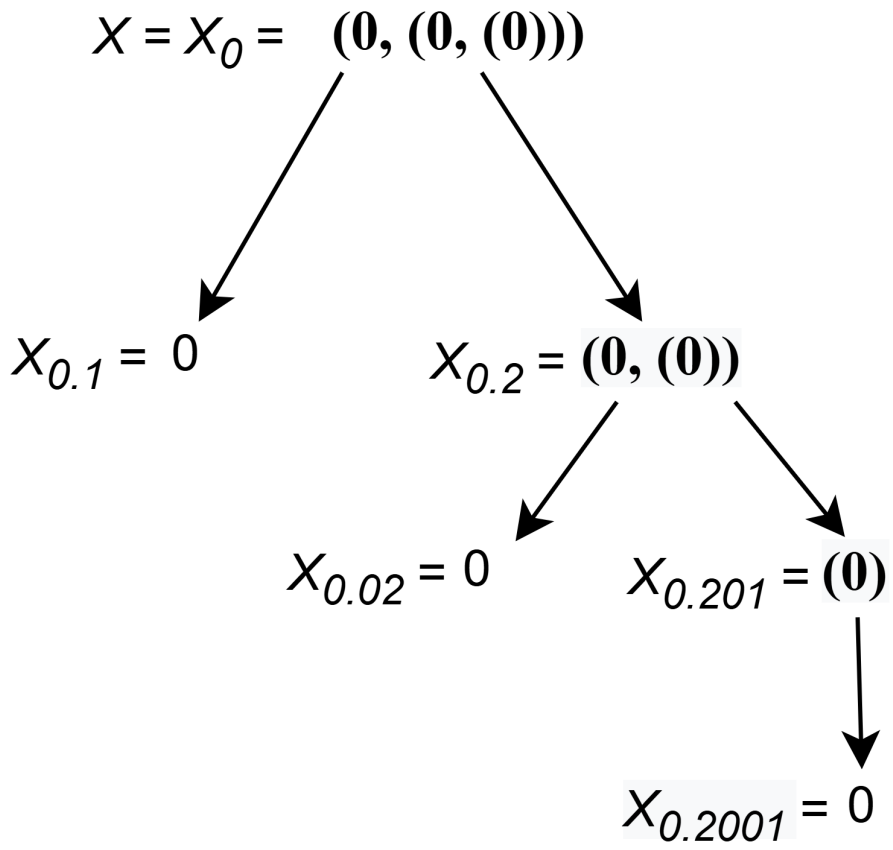

**Figure 6.** An example of the encoding of the nodes in the rooted tree structure of an E-formula.

Referring to the tree, that is, a connected graph without circuits, we call the tree vertex nodes. We elect a node as the starting point of each admissible itinerary and call it the root. We say that a node V'' is a child of a node V' if and only if every itinerary leading to V'' has V' as its last crossed node. We call branches the edges connecting the nodes. In order to give a sequential order to child nodes, we suppose that:

1. The graph of a tree is entirely drawn in a semi-plan originating from a straight line passing through the root;
2. All children of the same node are placed on the same straight line;
3. All straight lines so constituted are parallel to the one passing through the root; we call them rows and number, assigning 0 to the line passing through the root, and progressively counting the others starting from the 0 row;

**4.** The graph is observed by placing the straight line passing through the root parallel to the horizon of the observer's sight.

The children of a given node are listed proceeding from left to right. The first child is called the first-born and others, second-borns. Let now be given the set D of the rational numbers with finite decimals, where for each of its d elements is $0 \leq d < 1$. Then let's use the following conventions.

a. For every d of D, we will omit and make implicit the separator point for convenience. Thus, for example, 0032010431 stands for 0.032010431;

b. We will never denote an element of D with a decimal number ending with the digit 0. For example, to express 0.03, the strings 0030, 00300, etc., will never be used, as only 003 is admitted.

c. For each d belonging to D, we call the *terminal part* the sequence of digits following the last occurrence of the digit 0 in d. For example, if d = 0032010431, then the terminal part is 431, if d = 014, it is 14. On the other hand, we call the *initial part* the sequence of digits preceding the terminal part. In the first example, the initial part is 0032010 and in the second example, the initial part is 0.

d. We will call H the ordered set of all natural numbers where digit 0 never occurs, that is H = <1, ..., 9, 11, ...19, 21, ..., 99, 111, ..., 121, …>;

e. *n*[H] indicates the element occupying the *n*-th position in H.

**Definition** [Definition of the relation *f*]

We call *f* the relation assigning at each node of a tree T an element in D, in the following way:

- 0 is assigned to the root;
- proceeding in an orderly manner from left to right on row 1, to each node is assigned a number *d* of D with 0 as initial part, and respectively, numbers *1*[H], *2*[H], ... as terminal part [i.e, proceeding from left to right, to the nodes of this row are respectively assigned the following numbers of D: 01, 02, …, 09, 011, …, 099, 0111, 0112, …];
- Let V be any node of T, different from the root, having a correspondence with an element *d* of D according to relation *f*. If *d* = *pt*, where *p* is the initial part and *t* is the terminal part, then:
  - to the first-born V is assigned the number consisting of the string *p*0*t* [for example, if *d* = 012004, to the first-born is assigned 0120004];
  - to the second-born V are assigned numbers having the string pt0 as initial part, and, as terminal part, put in the following order the numbers *1*[H], *2*[H], … [in the previous example, being d = 012004, to the second-borns are assigned in the following order numbers 01200401, 01200402,…, 01200409, 0104011,…];

1. a node in the tree T is associated, according to the *f*, to an element of D if and only if it follows from 1)-3).

**Theorem 1.** [Relation *f* is a function]

For each tree T, relation *f* is a function going from the nodes of T to the set D.

**Proof**

We will proceed by induction on the number *i* of the *i*-th row of the tree T:

1. if $i = 0$, the row has only one node, the root; by the definition of the *f* relation, only the value 0 can be assigned to it;
2. if $i = 1$, the nodes of the *i* row are all children of the root: proceeding orderly from left to right, the unique number $d = 0n[H]$ is assigned to *n*-th child, where $n[H]$ is the *n*-th element of the H set.
3. let $i > 1$ and let V be any node of this row. For the inductive hypothesis, one and only one number $d = pt$ corresponds to it. According to the definition of the *f* relation, $p0t$ is thus assigned to the first-born of V. No other number can be assigned to this first-born. In fact, let's assume that a $p'0t'$ different from $p0t$ is assigned to this first-born. Then, two different numbers $d = pt$ and $d' = p't'$ should correspond to V against the inductive hypothesis. The proof proceeds in a similar way to all second-borns from V, to which are neatly assigned the numbers $pt01[H]$, $pt02[H]$, etc. In fact, the only possibility of non-univocally assignment to the second-born should concern only the *pt* part of these numbers. But this would imply that two different numbers, $d = pt$ and $d' = p't'$, are assigned to V, against the inductive hypothesis.

**Theorem 2** [The first-borns of a row $i > 1$ have more than one 0 immediately before the terminal part]
Let a tree T be given, and let's call $D_T$ the subset of D which is the codomain of $f[T]$. Then, an element of D whose initial part ends with more than one occurrence of digit 0 belongs to $D_T$ if and only if its counterimage is a first-born of a node belonging to a row $i > 1$.
**Proof**
Let $d = f[V] = pt$, where *p* ends with at least 2 occurrences of digit 0. By the definition of *f*, T can neither be the root of T nor belong to its row 1. Let's now assume ad absurdum that V is the second-born of some V' belonging to a row $i > 1$. Let $d' = f[V'] = p't'$. Then, for some $n[H]$, V must be $d = f[V] = p't'0\ n[H]$, that is, it should be $p = p't'0$. But in this case, *p* would end with a single occurrence of 0, against the hypothesis that it must end with at least 2 occurrences of 0.

**Theorem 3** [Function *f* is one-one]
For each T tree, calling $D_T$ the codomain of *f* in D, to each element of $D_T$ corresponds one and only one counterimage in T.
**Proof**
Let's now prove that function *f* is a one-one function, i.e., that according to *f*, one and only one counterimage in T corresponds to each element of $D_T$. The proof is carried out by induction on the number of occurrences of digit 0 in d belonging to $D_T$.

1. If $d = 0$, then its counterimage, by the definition of f, can only be the root of T.
2. If $d \neq 0$ and it has just one occurrence of digit 0, then, because of the definition of the terminal part of *d*, *d* must be equal to $0t$, where *t* is the terminal part. By the definition of *f*, the node corresponding to $d = 0t$ can only belong to row 1, and occupies exactly the same and unique position as *t* in H.
3. If *d* has two occurrences of digit 0, then it must be either $d = 00t$ or $d = 0q0t$, where *q* is a string with no occurrences of 0. In the first case, for Theorem 2 of, V is the first-born of V' such that $d' = f[V'] = 0t$. As seen at item 2, V' is unique, and its first-born V must also be unique. In the second case, for Theorem 2 of, $d = f[V] = 0q0t$ is only possible if V is a second born of some V' from row 1. By the definition of *f*, $d' = f[V'] = 0q$, and as for item 2, this V' is unique. Since $d = f[V] = 0q0t$, V is the node in the second-born sequence of V' occupying exactly the same and unique position as *t* in H. Thus, V is also unique.

4. If *d* has *i* occurrences of digit 0, where $i > 2$, then in $d = qzt$, where *q* ends with a digit different from 0, and *z* which is a string of 0 and *t* as the terminal part, we can have two cases: *z* is either constituted by more than one occurrence of 0, or it is constituted by only one occurrence of 0. In the first case, for Theorem 2, V is the first-born of some V' of the T tree. However, by the definition of *f*, d' = *f*[V'] = *qz't*, and *z'* has one less occurrence of 0. Therefore, by the inductive hypothesis, V' is unique, and consequently, so is V. Let now deal with the second case, the one in which *z* is constituted by a single occurrence of 0, that is, $d = q0t$. By Theorem 2 is a second born of some V', and by the definition of *f*, *d'* = *f*[V'] = q. Since d' has one occurrence of 0 less than d, by the inductive hypothesis, V' is unique. Therefore, V is also unique. In fact, since *d* = *f*[V] = *q*0*t*, V is precisely the node in the succession of the second-born of V' occupying exactly the same and unique position as *t* in H.

## References

Barsalou, L. W. {2020). Challenges and Opportunities for Grounding Cognition. *Journal of Cognition,* 3(1): 31, pp. 1–24.

Beer, R. D. (2000). Dynamical approaches to cognitive science. *Trends in Cognitive Sciences*, 4(3), 91–99.

Bender, E. M., Gebru, T., McMillan-Major, A., & Shmitchell, S. (2021). On the dangers of stochastic parrots: Can language models be too big? In *Proceedings of the 2021 ACM Conference on Fairness, Accountability, and Transparency* (pp. 610–623).

Bommasani, R., Hudson, D. A., Adeli, E., et al. (2021). On the opportunities and risks of foundation models. *arXiv preprint arXiv:2108.07258*.

Bostrom, N. (2014). *Superintelligence: Paths, Dangers, Strategies*. Oxford University Press.

Brooks, R. A. (1991). Intelligence without representation. *Artificial Intelligence*, 47(1–3), 139–159.

Brown, T. B., Mann, B., Ryder, N., et al. (2020). Language models are few-shot learners. *arXiv preprint arXiv:2005.14165v4.*

Bubeck, S., Chandrasekaran, V., Eldan, R., Gehrke, J., Horvitz, E., Kamar, E., Lee, P., Lee, Y. T., Li, Y., Lundberg, S., Nori, H., Palangi, H., Ribeiro, M. T., & Zhang, Y. (2023). Sparks of artificial general intelligence: Early experiments with GPT-4. *arXiv preprint arXiv:2303.12712*.

Buzsáki, G. (2019). *The Brain from Inside Out*. Oxford University Press.

Diba, K. & Buzsáki, G. (2007). Forward and reverse hippocampal place-cell sequences during ripples. Nat Neurosci 10, 1241–1242. doi: https://doi.org/10.1038/nn1961

Chollet, F. (2019). On the measure of intelligence. *arXiv preprint arXiv:1911.01547*.

Di Paolo, E. A., Buhrmann, T., & Barandiaran, X. E. (2017). *Sensorimotor Life: An Enactive Proposal*. Oxford University Press.

Floridi, L., & Chiriatti, M. (2020). GPT-3: Its nature, scope, limits, and consequences. *Minds and Machines*, 30, 681–694.

Goertzel, B. (2014). Artificial General Intelligence: Concept, State of the Art, and Future Prospects. *Journal of Artificial General Intelligence,* 5(1), 1- 48.

Harnad, S. (1990). The symbol grounding problem. *Physica D: Nonlinear Phenomena*, 42(1–3), 335–346.

Kelso, J. A. S. (1995). *Dynamic Patterns: The Self-Organization of Brain and Behavior*. MIT Press.

Lake, B. M., Ullman, T. D., Tenenbaum, J. B., & Gershman, S. J. (2017). Building machines that learn and think like people. *Behavioral and Brain Sciences*, 40, e253.

Legg, S., & Hutter, M. (2007). Universal intelligence: A definition of machine intelligence. *Minds and Machines*, 17(4), 391–444.

Marcus, G. (2020). The next decade in AI: Four steps towards robust artificial intelligence. *arXiv preprint arXiv:2002.06177*.

Maturana, H. R., & Varela, F. J. (1980). *Autopoiesis and Cognition: The Realization of the Living*. D. Reidel Publishing Company.

Morris, M. R., Sohl-Dickstein, J., Fiedel, N., Warkentin, T., Dafoe, A., Faust, A., ... & Legg, S. (2024). Levels of AGI: Operationalizing Progress on the Path to AGI. *arXiv preprint arXiv:2311.02462*.

Muñoz-Cristi, I. (2025). Steps Toward Anthroporoboticsbuilding: A Conversation With Humberto Maturana. *Constructivist Foundations,* 21(1), 1-12.

Newell, A., & Simon, H. A. (1976). Computer science as empirical inquiry: Symbols and search. *Communications of the ACM*, 19(3), 113–126.

Russell, S., & Norvig, P. (2021). *Artificial Intelligence: A Modern Approach* (4th ed.). Pearson.

Sarikaya, R. (2025). Path to Artificial General Intelligence: Past, present, and future. *Annual Reviews in Control*, 60, 101021.

Searle, J. R. (1980). Minds, brains, and programs. *Behavioral and Brain Sciences*, 3(3), 417–457.

Varela, F. J., Thompson, E., & Rosch, E. (1991). *The Embodied Mind: Cognitive Science and Human Experience*. MIT Press.

Wang, P. (2019). *On Defining Artificial Intelligence*. Journal of Artificial General Intelligence, 10(2), 1–37.

Wang, Y. and Sun, A. (2025). Toward Embodied AGI: A Review of Embodied AI and the Road Ahead. *arXiv preprint arXiv:2505.14235v1*.